\documentclass[10pt]{article}

\usepackage[margin=1in]{geometry}
\usepackage{amsmath}
\usepackage{amssymb}
\usepackage{graphicx}
\usepackage{multirow}
\usepackage{url}
\usepackage{subcaption}
\usepackage{caption}
\usepackage{pdflscape}
\usepackage{tikz}
\usetikzlibrary{fit,positioning}

\usepackage{algorithm} 
\usepackage{algorithmic} 

\newcommand*\samethanks[1][\value{footnote}]{\footnotemark[#1]}

\title{Blind Image Denoising via Dependent Dirichlet Process Tree}

\author{Fengyuan~Zhu \thanks{These authors contributed equally to this work}   \thanks{These authors are with the Department
		of Computer Science and Engineering, The Chinese University of Hong Kong. The contact is in http://www.cse.cuhk.edu.hk/~pheng/}
  \and   Guangyong~Chen \samethanks[1] \samethanks[2]
   \and  Jianye~Hao \thanks{Jianye Hao is with Tianjin University, China.}
	\and~Pheng-Ann~Heng \samethanks[2]}

\begin{document}

\maketitle
\begin{abstract}
Most existing image denoising approaches assumed the noise to be homogeneous white Gaussian distributed with known intensity. However, in real noisy images, the noise models are usually unknown beforehand and can be much more complex. This paper addresses this problem and proposes a novel blind image denoising algorithm to recover the clean image from noisy one with the unknown noise model. To model the empirical noise of an image, our method introduces the mixture of Gaussian distribution, which is flexible enough to approximate different continuous distributions. The problem of blind image denoising is reformulated as a learning problem. The procedure is to first build a two-layer structural model for noisy patches and consider the clean ones as latent variable. To control the complexity of the noisy patch model, this work proposes a novel Bayesian nonparametric prior called ``Dependent Dirichlet Process Tree'' to build the model. Then, this study derives a variational inference algorithm to estimate model parameters and recover clean patches. We apply our method on synthesis and real noisy images with different noise models. Comparing with previous approaches, ours achieves better performance. The experimental results indicate the efficiency of the proposed algorithm to cope with practical image denoising tasks.
\end{abstract}

{\section{Introduction}}
\label{sec_Intro}
{I}{mage} denoising is a fundamental problem which has been studied for decades in the area of computer vision and image processing~\cite{dabov2007image,elad2006image,buades2005non,lebrun2013nonlocal,wang2013sure,yu2012solving,dong2011sparsity,rajwade2013image}. Most of the previous approaches were developed under the assumption that the noise follows homogeneous white Gaussian distribution with fixed, known standard deviation $ \sigma $. However, the noise models of real images can be much more complex rather than one-parameter homogeneous white Gaussian noise. As stated by Tsin~\cite{tsin2001statistical}, in real images, the noise can be introduced by multiple different sources~(e.g., capturing instruments, data transmission media, image quantization and discrete source of radiation). The noise types can neither be Gaussian nor homogeneous especially when the series and band of the capture device, its setting~(ISO, aperture, shutter speed), as well as the image acquiring environment are unknown~\cite{liu2008automatic,lebrun2014multiscale}. As a result, the statistics of the noise can be signal, frequency, scale and spatial dependent. Thus, for practical use, image denoising algorithms must be flexible enough to efficiently cope with complex noise, even when the noise model is not provided. Such problem is defined as the ``blind image denoising''. 

Portilla~\cite{portilla2004blind,portilla2004full} proposed a generalized version of the BLS-GSM denoising method~\cite{portilla2003image}. This method relaxed the the assumption that the noise is homogeneous white Gaussian distributed by adopting a zero-mean correlated Gaussian model to estimate the noise for each wavelet subband. Liu et al.\cite{liu2008automatic} proposed a segmentation-based algorithm for JPEG image blind denoising to cope with intensity-dependent noise. The method first segmented an input image into small piecewise areas; then introduced a so-called ``noise level function''~(NLF) to model the relation between the gray level of a pixel and its noise level. A Gaussian conditional random field is constructed for denoising. Recently, Lebrun~\cite{lebrun2014multiscale} introduced a new approach called ``multiscale noise clinic'' with the state-of-the-art performance. It is an adaption of the NL-Bayes~\cite{lebrun2013nonlocal}. Similar to \cite{portilla2004blind} and \cite{portilla2004full}, this method introduced a zero-mean correlated Gaussian noise model for each group of similar patches and estimated the covariance matrix of noise for each group with the noise estimation algorithm in \cite{colom2014non}. 
\begin{figure}[t]
	\centering
	\includegraphics[width=0.8\textwidth]{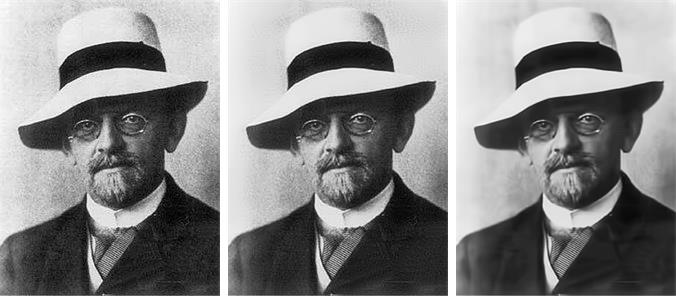}
	\caption{Left to right: noisy image, denoising results of \cite{lebrun2014multiscale} and ours.}
	\label{fig_hilbert}
\end{figure}

All these above algorithms made good efforts in investigating the relation between signal and noise. However, one common assumption they rely on is that the noise on certain group of patches or pixels within an image is homogeneous Gaussian distributed when the patches share certain features~(e.g., similar pixel intensity~\cite{liu2008automatic}, similar patches in transform domain~\cite{portilla2004full,portilla2004blind} or feature domain~\cite{lebrun2014multiscale}). These methods proposed different schemes to group patches or pixels in different domains from the noisy image. Then the noise is eliminated with the conjunction of a thorough noise estimation method for each group followed by an adapted denoising method. These methods achieved good performance, but the researchers also suggested that their noise models may still be inflexible for noise on real images as the empirical image noise can be more complex~\cite{tsin2001statistical,liu2008automatic,lebrun2014multiscale}. To cope with real noisy images, the following issues must be considered as well:
\begin{itemize}
	\item The Noise on real images can be frequency, intensity, scale and spatial dependent. The dependency relation between noise and the image is not fixed and can be very complex to be well modeled. 
	\item Even the dependency relation between noise and signal were known, resulting in the patches~(or pixels) being well grouped, the underlying noise model within each group may not be well estimated by Gaussian distribution~(e.g., the true noise model can be non-Gaussian and even multi-model in real cases). 
\end{itemize}
The noise model of these existing algorithms may still not be flexible enough to model the dependency between noise and signal for real noisy images. Fig.~\ref{fig_hilbert} shows an example of the denoising result of \cite{lebrun2014multiscale} on an old photo of David Hilbert~(downloaded from his Wikipedia profile\footnote{\url{https://en.wikipedia.org/wiki/David_Hilbert}}). The algorithm in \cite{lebrun2014multiscale} can eliminate some but not complete noise because of the loss of generality of its noise model.

We consider the blind image denoising from a new perspective. Instead of investigating the complex relation between noise and different image features, we directly model the empirical noise of an image with a multi-modal and non-Gaussian distribution which is flexible enough to cover a wide varieties of image noise models. A natural selection of such distributions is the mixture of Gaussian distribution~(MoG) which is not only a universal approximator to all continuous distributions, but also a fundamental multi-modal model for heterogeneous data~\cite{bishop2006pattern,meng2013robust,zhao2014robust}. To recover the clean image, we formulate the blind image denoising as a learning problem to estimate noisy patch model with complex noise~(MoG). We treat the clean image patches as latent variables~(or missing data) and estimate them when the model is learned. Motivated by the facts that clean natural image patches are well modeled by the Mixture of Gaussian distribution~\cite{zoran2012natural,theis2011all} and lie in local subspaces~\cite{gu2014weighted,soltanolkotabi2014robust}, we introduce a low-rank Gaussian mixture model for the modeling of the underlying clean image patches. A two-layer structural mixture model for observed noisy patches is derived accordingly.

To infer the free parameters within this model~(e.g., the number of components of the MoG for noise and clean patch model, the rank of each Gaussian component), this study introduces Bayesian nonparametric techniques for model construction. We propose a novel nonparametric prior called ``\textit{Dependent Dirichlet Process Tree}'' as a prior for the structural mixture model. We derive a novel nonparametric structural mixture model and approximated the full posterior distribution of the model with variational inference. Fig.~\ref{fig_hilbert} shows an example of the denoising result of our method comparing with \cite{lebrun2014multiscale}. The proposed approach efficiently eliminate the complex noise on the real images and well-preserve the detailed features.  

To summarize, the contribution of this work is five-fold as follows:
\begin{itemize}
	\item[1.] We introduce MoG to model the complex noise on real images, which is a new attempt in the area of image denoising.
	\item[2.] We formulate the blind image denoising in a new framework: first, build model for noisy patches and treat clean ones as latent variables; second, estimate the parameters of the model as well as the latent variables to recover the clean patches. 
	\item[3.] We propose a novel dependent nonparametric prior called ``\textit{Dependent Dirichlet Process Tree}'' and use this prior to build nonparametric Bayesian model for noisy patches.
	\item[4.] We develop a variational Bayesian algorithm to infer the proposed model to estimate the model as well as the latent clean patches for practical image denoising problems.
	\item[5.] We conduct extensive experiments on both synthetic and real noisy images with our method. The experimental results show that the proposed noise model is efficient for empirical noise on real-world images and the proposed method can cope with empirical image denoising tasks for real noisy images with the best performance.
\end{itemize}

The remainder of the paper is organized as follows: Sec.~\ref{sec_DPMM} introduces the background of Dirichlet process which is used to build the nonparametric model for blind image denoising. Sec.~\ref{sec_Model} introduces our novel framework for blind image denoising. Sec.~\ref{subsec_DDPT} derives an elegant nonparametric prior and Sec.~\ref{subsec_BID} applies it to build a Bayesian nonparametric model for blind image denoising. Sec.~\ref{sec_Inference} presents the derivation of posterior inference algorithms to learn the clean patches from noisy observations. Sec.~\ref{sec_Experiment} conducts extensive experiments of our method compared with related works. Sec.~\ref{sec_Conclu} provides the discussion of this study and future work.
\begin{figure*}[htp!]
	\centering  
	\begin{subfigure}[b]{0.3\textwidth}
		\centering
		\begin{tikzpicture}
		\tikzstyle{point1}=[circle, inner sep= 0.8mm,fill = black!100, draw =black!80]
		\tikzstyle{point2}=[circle, inner sep= 0.5mm,fill = black!10, draw =black!10]
		\tikzstyle{connect1}=[draw =black!80, thick]
		\tikzstyle{connect2}=[draw =black!10, thick]
		
		\node[point1] (l0) at (0,0) {} ;
		
		\node[point2] (l11) at (-1.2,-1) {};
		\node[point1] (l12) at (-0.4,-1) {};
		\node[point2] (l13) at (0.4,-1) {};
		\node[point2] (l14) at (1.2,-1) {};
		
		\node[point2] (l20) at (-1.8,-2) {} ;
		\node[point2] (l21) at (-1.4,-2) {};
		\node[point1] (l22) at (-0.9,-2) {};
		\node[point2] (l23) at (-0.5,-2) {};
		\node[point2] (l24) at (0,-2) {};
		\node[point2] (l25) at (0.4,-2) {};
		\node[point2] (l26) at (0.8,-2) {};
		\node[point2] (l27) at (1.4,-2) {} ;
		
		\draw[connect2] (l0) -- (l11);
		\draw[connect1] (l0) -- (l12);
		\draw[connect2] (l0) -- (l13);
		\draw[connect2] (l0) -- (l14);
		\draw[connect2] (l11) -- (l20);
		\draw[connect2] (l11) -- (l21);
		\draw[connect1] (l12) -- (l22);
		\draw[connect2] (l12) -- (l23);
		\draw[connect2] (l13) -- (l24);
		\draw[connect2] (l13) -- (l25);
		\draw[connect2] (l13) -- (l26);
		\draw[connect2] (l14) -- (l27);
		\draw[densely dotted, gray, very thin] (-1.7,-1.25) rectangle (2.3,-0.75);
		\draw[densely dotted, gray, very thin] (-2.3,-2.25) rectangle (2.9,-1.75);
		\node at (1.85,-1.05) {$\theta_1$};
		\node at (2.20,-2.05) {$\theta_1, \theta_2,\theta_3$};
		\end{tikzpicture}
		\caption{Component 1}
	\end{subfigure}  
		\begin{subfigure}[b]{0.3\textwidth}
			\centering
			\begin{tikzpicture}
			\tikzstyle{point1}=[circle, inner sep= 0.8mm,fill = black!100, draw =black!80]
			\tikzstyle{point2}=[circle, inner sep= 0.5mm,fill = black!10, draw =black!10]
			\tikzstyle{connect1}=[draw =black!80, thick]
			\tikzstyle{connect2}=[draw =black!10, thick]
			
			\node[point1] (l0) at (0,0) {} ;
			
			\node[point2] (l11) at (-1.2,-1) {};
			\node[point2] (l12) at (-0.4,-1) {};
			\node[point1] (l13) at (0.4,-1) {};
			\node[point2] (l14) at (1.2,-1) {};
			
			\node[point2] (l20) at (-1.8,-2) {} ;
			\node[point2] (l21) at (-1.4,-2) {};
			\node[point2] (l22) at (-0.9,-2) {};
			\node[point2] (l23) at (-0.5,-2) {};
			\node[point1] (l24) at (0,-2) {};
			\node[point2] (l25) at (0.4,-2) {};
			\node[point2] (l26) at (0.8,-2) {};
			\node[point2] (l27) at (1.4,-2) {} ;
			
			\draw[connect2] (l0) -- (l11);
			\draw[connect2] (l0) -- (l12);
			\draw[connect1] (l0) -- (l13);
			\draw[connect2] (l0) -- (l14);
			\draw[connect2] (l11) -- (l20);
			\draw[connect2] (l11) -- (l21);
			\draw[connect2] (l12) -- (l22);
			\draw[connect2] (l12) -- (l23);
			\draw[connect1] (l13) -- (l24);
			\draw[connect2] (l13) -- (l25);
			\draw[connect2] (l13) -- (l26);
			\draw[connect2] (l14) -- (l27);
			\draw[densely dotted, gray, very thin] (-1.7,-1.25) rectangle (2.3,-0.75);
			\draw[densely dotted, gray, very thin] (-2.3,-2.25) rectangle (2.9,-1.75);
			\node at (1.85,-1.05) {$\theta_1$};
			\node at (2.20,-2.05) {$\theta_1, \theta_2, \theta_3$};
			\end{tikzpicture}
			\caption{Component 2}
		\end{subfigure}  
			\begin{subfigure}[b]{0.3\textwidth}
				\centering
				\begin{tikzpicture}
				\tikzstyle{point1}=[circle, inner sep= 0.8mm,fill = black!100, draw =black!80]
				\tikzstyle{point2}=[circle, inner sep= 0.5mm,fill = black!10, draw =black!10]
				\tikzstyle{connect1}=[draw =black!80, thick]
				\tikzstyle{connect2}=[draw =black!10, thick]
				
				\node[point1] (l0) at (0,0) {} ;
				
				\node[point2] (l11) at (-1.2,-1) {};
				\node[point2] (l12) at (-0.4,-1) {};
				\node[point1] (l13) at (0.4,-1) {};
				\node[point2] (l14) at (1.2,-1) {};
				
				\node[point2] (l20) at (-1.8,-2) {} ;
				\node[point2] (l21) at (-1.4,-2) {};
				\node[point2] (l22) at (-0.9,-2) {};
				\node[point2] (l23) at (-0.5,-2) {};
				\node[point2] (l24) at (0,-2) {};
				\node[point2] (l25) at (0.4,-2) {};
				\node[point1] (l26) at (0.8,-2) {};
				\node[point2] (l27) at (1.4,-2) {} ;
				
				\draw[connect2] (l0) -- (l11);
				\draw[connect2] (l0) -- (l12);
				\draw[connect1] (l0) -- (l13);
				\draw[connect2] (l0) -- (l14);
				\draw[connect2] (l11) -- (l20);
				\draw[connect2] (l11) -- (l21);
				\draw[connect2] (l12) -- (l22);
				\draw[connect2] (l12) -- (l23);
				\draw[connect2] (l13) -- (l24);
				\draw[connect2] (l13) -- (l25);
				\draw[connect1] (l13) -- (l26);
				\draw[connect2] (l14) -- (l27);
				\draw[densely dotted, gray, very thin] (-1.7,-1.25) rectangle (2.3,-0.75);
				\draw[densely dotted, gray, very thin] (-2.3,-2.25) rectangle (2.9,-1.75);
				\node at (1.85,-1.05) {$\theta_1$};
				\node at (2.20,-2.05) {$\theta_1, \theta_2, \theta_3$};
				\end{tikzpicture}
				\caption{Component 3}
			\end{subfigure}  
			\caption{Three components of a tree-structured dependent mixture model. Each layer denotes a mixture model consist of certain components represented by nodes. In the bottom layer, each component is parameterized by $  \theta_1, \theta_2 $ and $ \theta_3 $. In the higher layer, each component is parameterized by $ \theta_1 $. The children of a node in the bottom layer shares the same $ \theta_2$ and $\theta_3 $, e.g., component 2 and 3. }
			\label{fig_tree}
\end{figure*}
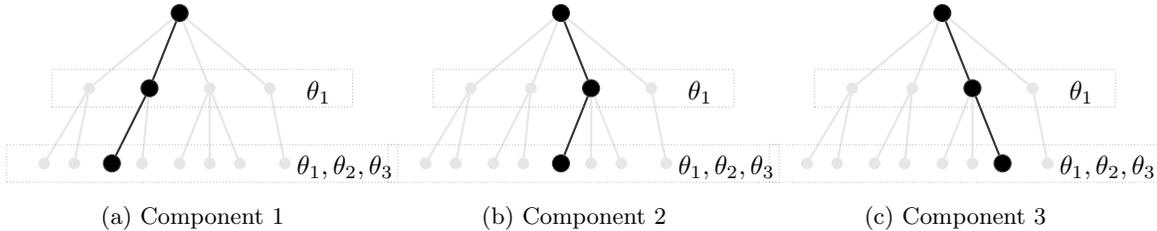
\section{Dirichlet Process}
\label{sec_DPMM}
The Dirichlet Process~\cite{ferguson1973bayesian} specifies a distribution over the space of probability measures on a measurable space~($ \Omega $, $ \mathcal{B} $), i.e., each draw of a Dirichlet Process is also a distribution. It is parameterized by the concentration parameter $ \alpha $ and base measurement $ H $. Let $ G $ be a sample drawn from $ DP(\alpha, H) $, Ferguson~\cite{ferguson1973bayesian} proved that, for all $ \{ B_1,\dots,B_k \} $ of $ \Omega $ with $ B_i \in \mathcal{B} $,
\begin{align*}
(G(B_1),\dots,G(B_k))\sim Dirichlet (\alpha H(B_1),\dots,\alpha H(B_k)).
\end{align*}
$ G $ is potentially infinite dimensional and almost sure discrete even when $ H $ is non-atomic~\cite{blackwell1973ferguson,sethuraman1991constructive}. As DP can generate discrete distributions on continuous parameter spaces, it is widely used as a prior for mixture models which is a linear superposition of component distributions~\cite{blei2006variational}. The basic form of a Dirichlet process mixture model is
\begin{align*}
G 			&\sim DP(\alpha, H)	\\
\theta_n 	&\sim G \\
x_n 		&\sim F(\cdot | \theta_n).
\end{align*} 
Given such representation, data $ x_1, \dots, x_N $ are considered to be drawn from distribution $ F $ with parameter $ \theta_1, \dots, \theta_N $. The parameters are drawn from the distribution $ G $. With a finite $ N $, the suitable dimension of $ G $, or number of components for the mixture model, can be learned to represent the data, as well as the parameters of the mixture.

The Dirichlet process mixture is a foundation for large number of Bayesian nonparametric models that rely on mixture representation of data. Even though $ G $ is abstract, it can be constructed by several methods for the convenience of inference. Next we review two of them to work with this infinite-dimensional distribution.

\subsection{Chinese Restaurant Process Construction}
\label{subsec_CRP}
The Chinese restaurant process~(CRP) is a simple constructive representation of DP~\cite{blackwell1973ferguson,aldous1985exchangeability}. Imagine that we have a Chinese restaurant with infinite many tables. Customers~(data) arrive and choose to sit at some table according to the following random process: 
\begin{itemize}
	\item[1.] The first customer sits at the first table with probability one, and orders a dish $\theta_1 \sim H$ 
	\item[2.] The $(N+1)$-th customer chooses the first unoccupied table with probability $\frac{\alpha}{N+\alpha}$, and orders a new dish~(parameter) $\theta_k \sim H$, or joins an occupied table with probability $\frac{c}{N+\alpha}$, where $ c $ is the number of people sitting at that table.
\end{itemize} 

Note that if $G \sim DP(\alpha,H)$ and $\theta_1,\ldots, \theta_n \sim G$, then $\theta_1,\ldots, \theta_n$ will follow a CRP. 
\subsection{Stick-breaking Construction}
\label{subsec_SBP}
The stick-breaking construction of DP is introduced by Sethuraman~\cite{sethuraman1991constructive}, which allows one to construct $ G $ directly before drawing $ \theta_n $. Consider two infinite collections of independent random variables, $ \theta_i \sim H$ and $ V_i \sim Beta(1, \alpha) $ for $ i = \{1,2,\dots\} $. If 
\begin{align*}
\pi_i &= V_i \prod_{j=1}^{i-1}(1-V_j)	\\
G     &= \sum_{i=1}^{\infty}\pi_i \delta_{\theta_i},
\end{align*}
we have $ G \sim DP(\alpha, H) $, where $ \delta_{\theta_i} $ is the Dirac probability measure concentrated at $ \theta_i $. The variable $ V_i $ can be considered as the proportion broken from the remainder of a unit-length stick with size proportional to random draws from $ Beta(1, \alpha) $. Then the weight $ \pi_i $ of a component $ i $ is the length of each of the infinite pieces of stick, $ V_i \prod_{j=1}^{i-1}(1-V_j) $. More random variables in the range of $[0,1]  $ are multiplied and thus the weights tends to zero exponentially. A nice feature of the procedure is that the construction of $ G $ is independent with $ \theta_1,\dots, \theta_N $, which is a significant advantage of this representation over CRP for mean-field variational inference.

\section{Proposed Model}
\label{sec_Model} 
Our goal is to develop a Bayesian nonparametric method for blind image denoising. We formulate the denoising problem as a learning procedure. We constructed a model for noisy patches and treated the clean ones as latent variables~(missing data) of the model. We estimate the model, as well as the latent variables with the observed noisy patches, and the clean images can be well-recovered accordingly. 

Let $ i \in \{1,\dots,N \} $ and $ N $ is the total number of data instances, an observed noisy patch $ x_i\in \mathcal{R}^d $, contaminated by noise $ e_i\in \mathcal{R}^d $, can be expressed as
\begin{align}
\label{Equation_denoise}
x_i = \hat{x}_i + e_i,
\end{align}
where $ \hat{x}_i\in \mathcal{R}^d $ is a latent variable corresponding to the underlying clean patch, $ d $ is the dimension of each patch. To generate $ x_i $, the procedure can be naturally split into two parts: (1) a clean patch $ \hat{x}_i $ is first drawn from a clean patch model as a latent variable, (2) $ x_i $ is drawn with $ \hat{x}_i $ and the noise model.  We identify two significant features of clean natural image patches to model the clean patches. First, as investigated in \cite{zoran2011learning,zoran2012natural}, clean natural image patches can be efficiently modeled by MoG. This attribute has also been used in other image processing algorithms~\cite{wang2013sure,yu2012solving,zoran2011learning}. Second, the real-world high dimensional signals~(including clean image patches) are always well-modeled by certain low-rank representation~\cite{vidal2010tutorial} since natural signals are rarely of full rank~\cite{vidal2010tutorial,wang2013sure}. Following the above features, we propose a mixture of low-rank Gaussian distribution~(low-rank MoG) with $ T $ components for clean patch modeling. The clean patches associated within the same component are assumed to lie on a subspace. The parameters associated with the low-rank MoG include a $ T $-dimensional probability vector $ \boldsymbol{\pi} = [ \pi_1 \dots \pi_T ] $ for component weights and $ \boldsymbol{\omega} = \{\omega_1,\dots,\omega_T\} $, where $ \omega_t $ is the parameters of the $ t $th subspace with $ t = 1,\dots,T $. However, the noise model is unknown for blind image denoising. For patches sharing the same subspace $ t $, we model the noise associated on them by a MoG with $ K_t $ components. The idea is reasonable as MoG is a universal approximator for all continuous distribution~\cite{bishop2006pattern}. Similar noise modeling strategy can also be found in \cite{meng2013robust} and \cite{zhao2014robust}. Each MoG is parameterized by the component weight vector $ \boldsymbol{\kappa_t} = [\kappa_{t,1} \dots \kappa_{t, K_t}] $ and $ \boldsymbol{\phi_t} = \{\phi_{t,1},\dots,\phi_{t,K_t}\} $, where $ \phi_{t,k} $ corresponds to the parameters of the $ k $th Gaussian component on the $ t $th subspace with $ k = 1,\dots,K_t $.

Under the above model, we can interpret the procedure of generating a noisy patch $ x_i $ as follows: first, a clean patch~(latent variable) is generated from a subspace with parameter $  \omega_t $; after that, $ e_i $ is drawn from a component of the MoG associated on subspace $ t $ with parameter $ \phi_{t,k} $. Thus, $ x_i $ can be considered to be generated from $ p(\cdot | \omega_t, \phi_{t,k}) = \int \text{d}\hat{x}_i p(\cdot |\phi_{t,k}, \hat{x}_i)p(\hat{x}_i | \omega_t) $. From this perspective, the noisy patch model is essentially a tree-structured dependent mixture model. At the top layer, the observed noisy patches are divided into $ T $ groups. At the second layer, each group is a mixture of components sharing the same $ \omega $, i.e., the latent clean patches of the noisy ones in the same group are drawn from the same subspace. 

In this section, we propose a statistical model for the above procedure as the noisy patch model. The clean patches are considered to be latent variables. As long as the above-mentioned hierarchies are explored, the parameters and the latent variables~(clean patches) can be well estimated. Finally, the clean image can be recovered accordingly. On building the model, we adopt Bayesian nonparametric methodology to inference the complexities of both layers~(e.g., $ T, K_1, \dots, K_t $) and build an elegant dependent Dirichlet process called \textit{Dependent Dirichlet Process Tree}~(DDPT) for this problem. DDPT can efficiently capture the dependency relation of parameters in a tree-structure. Then, we apply the DDPT on noisy patch modeling for the problem of blind image denoising. 

\subsection{Dependent Dirichlet Process Tree}
\label{subsec_DDPT}
The DP can be used to build nonparametric models that rely on mixtures to represent distribution on data. However, the process cannot capture the structure among the components. For a mixture model, where each component is parameterized by a distinct $ \Psi_L = \{\theta_{1}, \theta_{2},\dots,\theta_{L}\} $. In real cases, several components may share certain subset of $ \Psi_L $, denoted as $ \Psi_j = \{ \theta_1, \dots, \theta_{j} \} $. To capture this property, we divide the data into different groups and each group is represented by a mixture model whose components share the same $ \Psi_j $. Each group can be considered as a component of a higher layer mixture model, which is parametrized by $ \Psi_j = \{\theta_{1}, \theta_{2},\dots,\theta_{j}\} $. This kind of dependency can occur in the higher-layer mixture recurrently and form a hierarchical dependency tree among the mixture components with at most $ (L+1) $ layers~(the first layer has only one component with all data). Each component at the lowest-layer mixture model lies along a path in a hierarchy. An illustrating example is shown in Fig.~\ref{fig_tree} which may also explain a bit here.
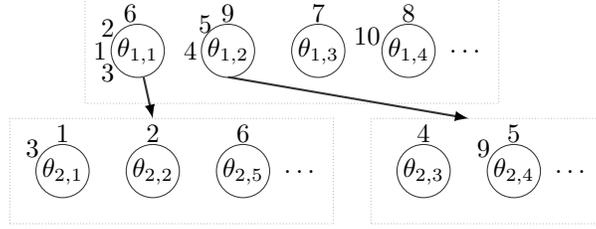
\begin{figure}[t]
	\centering
	\begin{tikzpicture}
	\tikzstyle{point1}=[circle, inner sep= 2.5 mm,fill = white!100, draw =black!80]
	\tikzstyle{connect1}=[-latex,draw =black!80, thick]

	\node[point1] (l11) at (-1.8,0) {};
	\node at (-1.8,0) {$\theta_{1,1}$};
	
	\node at (-2.3,0) {$1$};
	\node at (-2.2,0.3) {$2$};
	\node at (-2.2,-0.3) {$3$};
	\node at (-1.9,0.5) {$6$};
	
	\node[point1] (l12) at (-0.6,0) {};
	\node at (-0.6,0) {$\theta_{1,2}$};

	\node at (-1.1,0) {$4$};
	\node at (-0.9,0.35) {$5$};
	\node at (-0.6,0.5) {$9$};
	
	\node[point1] (l13) at (0.6,0) {};
	\node at (0.6,0) {$\theta_{1,3}$};
	\node at (0.6,0.5) {$7$};
	
	\node[point1] (l14) at (1.8,0) {};
	\node at (1.8,0) {$\theta_{1,4}$};
	\node at (1.8,0.5) {$8$};
	\node at (1.25,0.2) {$10$};
	
	\node at (2.55,0) {$\ldots$};
	
	\draw[densely dotted, gray, very thin] (-2.5,0.7) rectangle (3,-0.7);

	\node[point1] (l21) at (-2.8,-1.6) {};
	\node at (-2.8,-1.6) {$\theta_{2,1}$};
	\node at (-2.8,-1.1) {$1$};
	\node at (-3.2,-1.3) {$3$};
	
	\node[point1] (l22) at (-1.6,-1.6) {};
	\node at (-1.6,-1.6) {$\theta_{2,2}$};
	\node at (-1.6,-1.1) {$2$};
	
	\node[point1] (l25) at (-0.4,-1.6) {};
	\node at (-0.4,-1.6) {$\theta_{2,5}$};	
	\node at (-0.4,-1.1) {$6$};
	
	\node at (0.35,-1.6) {$\ldots$};
	\draw[densely dotted, gray, very thin] (-3.5,-2.3) rectangle (0.8,-0.9);

	\node[point1] (l23) at (2,-1.6) {};
	\node at (2,-1.6) {$\theta_{2,3}$};
	\node at (2,-1.1) {$4$};
	
	\node[point1] (l24) at (3.2,-1.6) {};
	\node at (3.2,-1.6) {$\theta_{2,4}$};
	\node at (3.2,-1.1) {$5$};
	\node at (2.8,-1.3) {$9$};
	
	\node at (3.95,-1.6) {$\ldots$};
	\draw[densely dotted, gray, very thin] (1.3,-2.3) rectangle (4.4,-0.9);
	
	\draw[connect1] (l11) -- (-1.6,-0.9);
	\draw[connect1] (-0.6,-0.35) -- (2.6,-0.9);
	 (1.9,-1.75);
	\end{tikzpicture}
	\caption{The representation of ``Chinese Restaurant Tourism Process''. Each layer corresponds to a city and the boxes are the Chinese restaurants in the city. Each circle is one of the infinite tables in each restaurant. A tourist~(represented by number) choose a table in a restaurant with CRP and the parameter on the table is the dish ordered.}
	\label{fig_CRP}
\end{figure}  

Here, we build the DDPT as a prior for structured multi-layer representation of mixture model to capture such dependency relationship. A \textit{Dependent Dirichlet Process Tree} with $ L $ layers can be defined by imaging the following scenario. Suppose that there are $ L $ cities along a travel route. The first city has one Chinese restaurant with infinite number of tables, and the rest of them all have infinite number of such Chinese restaurants. On each table in the restaurants of a city, there is a card with the name of another restaurant in the next city along the travel-line as a recommendation. Thus, the restaurants of all the cities are organized into a $ L $-layer tree with infinite branches.

A tourist arrives in the first city and begins the travel along the route. He plans to go to Chinese restaurant once in each of the cities. So he enters the only Chinese restaurant in the first city and selects a table according to a CRP in Sec.~\ref{subsec_CRP}. Then, he follows the recommendation on the table and goes to the restaurant identified on the table of the next city, again selects a table according to CRP. At the end of the trip, the tourist has visited exactly $ L $ restaurants which constitutes a path from the only restaurant of the first city. We name the above process the ``\textit{Chinese Restaurant Tourism Process}'' which can construct a DDPT and an illustrating example is shown in  Fig.~\ref{fig_CRP}. After $ N $ tourists finish the same travel schedule, a $ L $-level tree with potentially infinite branches is built with the collection of the tourists' paths.

\subsubsection{DDPT for Dependent Mixture Model}
The DDPT can be used as a prior to model the hierarchies of the dependent mixture model discussed before. For instance, consider a MoG whose components are Gaussian distributed. Each component $ i $ is parameterized by $ \mu_i $ and $ \Sigma_i $ as the mean vector and covariance matrix respectively. Here, we define the \textit{dependent MoG} as follows: the top layer represents a mixture of MoG~(MMoG). The Gaussian components of each MMoG component share the same mean vector. The second layer is formed by the Gaussian components of these MoG. 

A two-layer DDPT can be used as a prior to build a nonparametric dependent MoG. For each data $ x_t $, the mean vector $ \mu_t $ is drawn from  a CRP as Sec.~\ref{subsec_CRP} at first. Then $ \Sigma_t $ is drawn from the CRP that $ \mu_t $ identifies. The generative model of this procedure is
\begin{equation*}
\begin{split}
\begin{matrix}
& G_1 \sim \text{DP}(\alpha_1, H_1) 				&\mu_t \sim	G_1\\
& G_{2, \mu_t}\sim   \text{DP}(\alpha_2, H_2) 	&\Sigma_t	\sim	G_{2, \mu_t}\\
\end{matrix}\\
 x_t			\sim	p(\cdot | \mu_t, \Sigma_t).\hspace{14mm}
\end{split}
\end{equation*}
Here, $ \mu_t $ is drawn from the first layer DP with concentration parameter $ \alpha_1 $ and base measurement $ H_1 $.  $ \Sigma_t $ is drawn from the second layer DP of the sample's group with concentration parameter $ \alpha_2 $ and base measurement $ H_2 $. 

By repeating the above process, it is straightforward to derive the extension of the above generative model for DDPT with more layers. 
\subsubsection{Stick-breaking Construction for DDPT}
\label{subsec_SBP_ours}
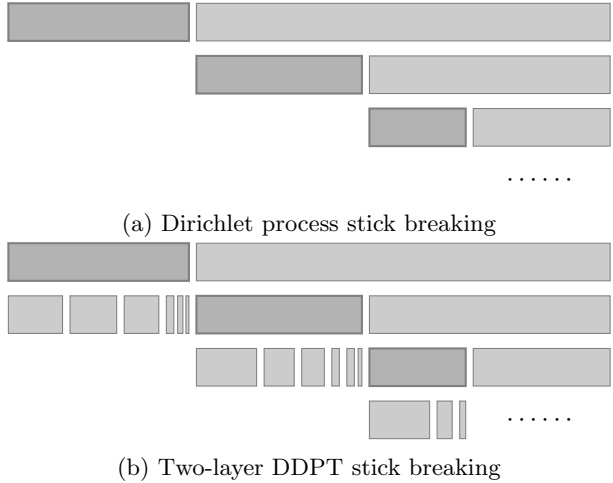
\begin{figure}[htp!]
	\centering  
	\begin{subfigure}[b]{0.5\textwidth}
		\centering
		\begin{tikzpicture}
		\tikzstyle{box1}=[fill= gray!60!white, draw= gray!100!white,thick]
		\tikzstyle{box2}=[fill= gray!40!white, draw= gray!100!white]
		\filldraw[box1] (-4,-0.25) rectangle (-1.6,0.25);
		\filldraw[box2] (-1.5,-0.25) rectangle (4,0.25);
				
		\filldraw[box1] (-1.5,-0.95) rectangle (0.7,-0.45);
		\filldraw[box2] (0.8,-0.95) rectangle (4,-0.45);	
			
		\filldraw[box1] (0.8,-1.65) rectangle (2.08,-1.15);
		\filldraw[box2] (2.18,-1.65) rectangle (4,-1.15);
		\node at (3.09,-2.1) {$\cdots \cdots$};
		\end{tikzpicture}
		\caption{Dirichlet process stick breaking}
	\end{subfigure}  	
	\begin{subfigure}[b]{0.5\textwidth}
		\centering
		\begin{tikzpicture}
		\tikzstyle{box1}=[fill= gray!60!white, draw= gray!100!white,thick]
		\tikzstyle{box2}=[fill= gray!40!white, draw= gray!100!white]
		\filldraw[box1] (-4,-0.25) rectangle (-1.6,0.25);
		\filldraw[box2] (-1.5,-0.25) rectangle (4,0.25);

		\filldraw[box2]  (-4,-0.95) rectangle (-3.28,-0.45);
		\filldraw[box2]  (-3.18,-0.95) rectangle (-2.56,-0.45);
		\filldraw[box2]  (-2.46,-0.95) rectangle (-2,-0.45);
		\filldraw[box2] (-1.9,-0.95) rectangle (-1.8,-0.45);
		\filldraw[box2]  (-1.75,-0.95) rectangle (-1.68,-0.45);
		\filldraw[box2]  (-1.64,-0.95) rectangle (-1.6,-0.45);
		
		\filldraw[box1]  (-1.5,-0.95) rectangle (0.7,-0.45);
		\filldraw[box2]  (0.8,-0.95) rectangle (4,-0.45);

		\filldraw[box2]  (-1.5,-1.65) rectangle (-0.7,-1.15);
		\filldraw[box2]  (-0.6,-1.65) rectangle (-0.2,-1.15);
		\filldraw[box2]  (-0.1,-1.65) rectangle (0.2,-1.15);
		\filldraw[box2]  (0.3,-1.65) rectangle (0.4,-1.15);
		\filldraw[box2]  (0.5,-1.65) rectangle (0.6,-1.15);
		\filldraw[box2]  (0.65,-1.65) rectangle (0.7,-1.15);
		
		\filldraw[box1]  (0.8,-1.65) rectangle (2.08,-1.15);
		\filldraw[box2]  (2.18,-1.65) rectangle (4,-1.15);

		\filldraw[box2]  (0.8,-2.35) rectangle (1.6,-1.85);
		\filldraw[box2]  (1.7,-2.35) rectangle (1.9,-1.85);
		\filldraw[box2]  (2,-2.35) rectangle (2.08,-1.85);
		\node at (3.09,-2.1) {$\cdots \cdots$};
		\end{tikzpicture}
		\caption{Two-layer DDPT stick breaking}
	\end{subfigure}  	
	\caption{(a) Dirichlet process stick-breaking procedure, with a linear partitioning. (b) Two-layer DDPT stick breaking process. A stick with unit length is partitioned into stick segments via a stick-breaking process. Each stick segment is set to be unit length and an extra stick-breaking process is performed.}
	\label{fig_stickbreaking}
\end{figure}

The DDPT can also be represented with a stick-breaking construction. Let the root stick's length be $ \pi_0 = 1 $. At the first layer, the stick is broken via a stick-breaking process in Sec.~\ref{subsec_SBP} with parameter $ \alpha_0 $, i.e., the length of the $ i $th segment is $ \pi_{1,i} = \pi_0 V_{1,i}\prod_{j=1}^{i-1}(1-V_{1,j}) $ for $ i = \{ 1,2,\dots \} $. Then, at the second layer, the stick-breaking is applied to each of the stick segments at the first layer, e.g., for the $ k $th layer $ \pi_{1,k} $, the stick lengths of its $ i $th child segments is $ \pi_{1,k, i} = \pi_{1,k} V_{1,k,i}\prod_{j=1}^{i-1}(1-V_{1,k,j}) $ with $ i \in \{ 1,2,\cdots \} $. Such process continues for the $ L $ layers, which can be understood with Fig.~\ref{fig_stickbreaking}.
\subsubsection{Remarks on DDPT}
The proposed DDPT is not the only model to capture the dependency among mixture components in the area of Bayesian nonparametrics. Hierarchical DP~(HDP)~\cite{teh2006hierarchical} introduced a sharing mechanism to allow the sharing of mixture components for multiple Dirichlet process mixture models. The nested Chinese restaurant process~\cite{griffiths2004hierarchical,blei2010nested,wang2009variational} is a tree-structured extension of CRP for hierarchical clustering and the hierarchies built by nested CRP is similar with ours. But the dependency of parameters among components is not modeled explicitly. \cite{paisley2015nested} is the combination of the above two models to allow the sharing of mixture components for different sub-trees. \cite{yerebakan2014infinite} is an approach for modeling dependencies among mixture components, which is designed for mixture of MoG. It proposes DP with different base distributions for each MoG. And the base distributions are generated from a higher-layer DP. All these approaches achieve excellent performance under their consideration. But the problems they address focus on building the hierarchies for clustering, which is different from ours, resulting in the distinction between DDPT and previous approaches. 
 
\subsection{Two-layer DDPT for Blind Image Denoising}
\label{subsec_BID}
We apply the proposed DDPT on blind image denoising to model noisy patches and recover the underlying clean patches. A two-layer structural mixture model is built as follows. The patches are represented by $ T $ groups at the top layer to represent $ T $ subspaces and each group is represented by a mixture with $ K_t $ components for modeling noise. All the components of group $ t $ shares the same $ \omega_t $ with $ t\in\{ 1,\dots,T \} $ and the $ k $th component in group $ t $ is parameterized by $ \phi_{k,t} $. As long as these parameters are estimated, the clean patch model and noise model are recovered. Then, we can recover each clean patch by projecting the noisy one onto the clean patch model. 

We adopt DDPT as a prior to construct the structural mixture model and the generative model is 
\begin{equation}
\label{Eq_DDPT_denoise}
	\begin{split}
	\begin{matrix}
	& G \sim \text{DP}(\alpha, H) 				&\omega_i \sim	G\\
	& G_{\omega_i}\sim \text{DP}(\beta, G_0) 	&\phi_{i}	\sim G_{\omega_i}\\
	\end{matrix}\\
	x_i	 \sim	p(\cdot	|	\omega_i, \phi_i).\hspace{14mm}
	\end{split}
\end{equation}
In this model, the top layer DP generates $ \omega_i $ according to the base distribution $ H $ and the concentration parameter $ \alpha $. The samples sharing $ \omega_i $ are considered to be in the same group. Each group is represented as a DP mixture and the parameter $ \phi_i $ is drawn from the second layer DP of the group with base distribution $ G_0 $ and concentration parameter $ \beta $. Noisy patches can be drawn from this model with a stick-breaking construction as follows:
\begin{itemize}
	\item[1.] Draw $ v_t \sim Beta(1,\alpha) $, $ t = \{ 1,2,\cdots \} $ 
	\item[2.] Draw $ \omega_t \sim H $, $ t = \{ 1,2,\cdots \} $ 
	\item[3.] $ \pi_t = v_t \prod_{i=1}^{t-1}(1-v_i) $
	\item[4.] For each subspace $ t $
		\begin{itemize}
			\item[1.] Draw $ w_{t, k} \sim Beta(1, \beta) $, $ k = \{1,2,\cdots\} $
			\item[2.] Draw $ \phi_{t,k} \sim G_0 $, $ k = \{1,2,\cdots\} $
			\item[3.] $ \kappa_{t, k} = w_{t,k} \prod_{i=1}^{k-1}(1-w_{t,k}) $
		\end{itemize}
	\item[5.] For each data point $ x_i $ 
			\begin{itemize}
				\item[1.] Draw $ z_i $ from $ Mult(\boldsymbol{\pi}) $
				\item[2.] Draw $ \tilde{z}_i $ from $ Mult(\boldsymbol{\kappa}_{z_i}) $
				\item[3.] Draw $ x_i \sim p(\cdot | \omega_{z_i}, \phi_{z_i, \tilde{z}_i})$
			\end{itemize}
\end{itemize}

\subsubsection{Modeling Low-rank}
We propose low-rank MoG for the underlying clean patch modeling. Specifically, each clean patch~(latent variable) is assumed to follow a MoG, and the samples drawn from each Gaussian component lie in a local subspace. We model the low-rank property of each Gaussian component by enforcing its covariance matrix to be a low-rank positive semidefinite matrix. 

Given a $ d $-dimensional zero-mean low-rank Gaussian distribution $ \mathcal{G}(\mu,\Lambda) $, we decompose a sample $ x_i $ that drawn from it, with $ x_i = A y_i + \mu $. Here, $ A $ is a $ d \times d $ matrix while $ y_i $ is a random variable drawn from $ \mathcal{G}(0, I) $, a $ d $-dimensional zero-mean Gaussian distribution with identity covariance matrix. A similar decomposition method can also be seen in the area of dimensional reduction~\cite{tipping1999probabilistic,ghahramani1996algorithm}. With such decomposition, we have $ \Lambda = AA^T $. To model the low-rank feature of $ \mathcal{G}(\mu,\Lambda) $, an intuitive method is to enforce the matrix $ A $ to be low-rank since $ \text{rank}(A) = \text{rank}(AA^T) $. A nice attribute favored by this model is that the samples drawn from such model are also low-rank representable. Given n samples from $ \mathcal{G}(\mu,\Lambda) $ $ X =[x_1, x_2, \dots, x_n] $, $ X $ can be decomposed with $ X = AY + U$, where $ Y = [y_1, y_2, \dots, y_n] $, each $ y_i $ is random variable of $ \mathcal{G}(0, I) $ and $ U = [\mu, \dots,\mu] $ whose rank is 1. As $ \text{rank}(AB) \leq \min(\text{rank}(A), \text{rank}(B)) $ and $ \text{rank}(A+B) \leq \text{rank}(A) + \text{rank}(B) $, $ X $ is also low-rank since $ A $ is low-rank. Thus, all the samples generated by such model lie on a low-rank subspace.

A simple way to model the low-rank property of matrix $ A $ is to impose a trace norm prior~\cite{candes2009exact,recht2010guaranteed} over it. Following the setting in \cite{nakajima2013global,mnih2007probabilistic}, we propose the trace-norm prior on $ p(A) $ with
\begin{align}
\label{Eq_tracenorm}
p(A) \propto \exp(-\frac{1}{2}\text{Tr}(A C_A^{-1} A^T)).
\end{align}
Here $ \text{Tr}(\cdot) $ is the trace of a matrix. We assume that the prior covariance matrix $ C_A $ is diagonal and positive definite with
\begin{align*}
C_A = \text{diag}(c_{a_1}^2,\dots,c_{a_d}^2),
\end{align*}
for $ c_{a_h} > 0 $. Since $\text{Tr}(A C_A^{-1} A^T) = \text{Tr}(C_A^{-1} A^T A)$, we can obtain that $p(A)$ is proportion to $\prod_{j=1}^{d}\exp(-\frac{1}{2 c_{j}^2}\tilde{a}_{j}^T \tilde{a}_{j} )$. Thus, we have the detailed formulation of $p(A)$ as follows, 
	\begin{align*}
	p(A) 	= \prod_{j=1}^{d} \mathcal{G}(\tilde{a}_{j} | 0, c_{a_j}^2 {\bf I}),
	\end{align*}
where $ \tilde{a}_{j} $ denotes the $ j $th row vector of $ A $, which we assume to be independent with each other. $ \mathcal{G}(\tilde{a}_{j} | \mu, \Sigma) $ is Gaussian distribution with mean $ \mu $ and covariance matrix $ \Sigma $.

\subsubsection{Base Distributions}
We turn to discuss the choice of base distribution $ H $ and $ G_0 $ in Eq.~(\ref{Eq_DDPT_denoise}). The parameter of each subspace is denoted as $ \omega_t = \{ \mu_t, A_t  \} $, where $ \mu_t $ is the mean vector of subspace $ t $ and $ A_t $ represents the dictionary matrix of subspace $t$. For the base distribution $ H $, we set the parameters $\{\mu_t,A_t\}$ drawn according to the following generative process:
\begin{align*}
\mu_t		&\sim \mathcal{G}(\cdot| \mu_{0}, \Sigma_0),		\\
A_t	&\sim \frac{1}{\mbox{const}}\exp(-\frac{1}{2}\text{Tr}(A_t C_A^{-1} A_t^T)),
\end{align*}
where $ \text{const} $ is the normalization term. A components of the MoG noise associated on subspace $ t $ is parameterized by $ \phi_{t,k} = \{ u_{t,k}, \Upsilon_{t,k} \} $, where $ u_{t,k} $ is the mean vector and the covariance matrix is $ \Upsilon_{t,k} $ for each Gaussian component respectively, where $ D $ is the dimension of $ u_{t,k} $. Following the setting in \cite{corduneanu2001variational,blei2006variational}, we introduce the conjugate prior of Gaussian distribution as base distribution with the following generative process:
\begin{align*}
u_{t,k}  & \sim \mathcal{G}(\cdot|\varepsilon_0, \Omega_{0}), 	\\
\Upsilon_{t,k} & \sim \mbox{iWishart}(\nu_0, B_0),
\end{align*} 
where $ d \in \{1,\cdots,D\} $, the $ \mbox{iWishart}(\nu_0, B_0)$ is the inverse-Gamma distribution with shape $ \nu_0 $ and scale matrix $ B_0 $.

\subsection{Remarks on the Proposed Model}
In our approach, we assume that the clean patches lie on several subspaces and the noise associated on each subspace is MoG distributed due to MoG's efficiency in approximating continuous density. Thus, the marginal noise distribution of our approach is the convex combination of several MoGs, which remains to be a MoG. Following the results in \cite{tsin2001statistical,liu2008automatic,lebrun2014multiscale} that the mean of noise should be zero. We set the marginal mean of MoG noise of each subspace to be zero accordingly. 

In previous approaches, the noise is assumed to be dependent with signal. Here, we show that the noise model of these approaches can be interpreted by MoG with certain constraints as well, which further confirms that using MoG for noise modeling is reasonable. In \cite{portilla2004blind} and \cite{portilla2004full}, the noise on each wavelet subband is a Gaussian distribution. Thus, the marginal noise model is essentially a MoG with the constraint that the noise on each subband is drawn from single Gaussian component. Liu's approach~\cite{liu2008automatic} assume that the noise model is a density-dependent Gaussian distribution, which is also a MoG essentially. Lebrun's approach~\cite{lebrun2013nonlocal} assume that the noise of a patch, as well as its nearby patches in feature domain, is Gaussian distributed. The marginal noise distribution of a noisy image is MoG as well. In contrast, our noise model is more general. Without any assumption of the dependency relation between signal and noise, the algorithm estimates the empirical noise model from noisy patches directly. Specifically, the first layer DP mixture divides the noisy patches into groups with respect to the structure of their underlying clean patches. Within each group, a noisy patch can be considered as the addition of two independent signals, a low rank clean patch and a MoG noise. And the clean patch can be well recovered with Bayesian approaches. 

\section{Posterior Inference and Patch Recovery}
\label{sec_Inference}
In this section, we propose a variational approach to approximate the posterior distributions over latent variables and  parameters in the above model for image denoising. Given the observable data $X$, the Bayesian posterior distributions over latent parameters $\Theta$ can be represented as:
\begin{equation}
p(\Theta|X,\Delta) = \frac{p(X|\Theta,\Delta)p(\Theta|\Delta)}{p(X|\Delta)},
\label{Eq_VBPoster}
\end{equation}
where $\Delta$ denotes the hyperparameters of our DDPT model~(parameters of prior distributions). However, the analytical form of the posterior $p(\Theta|X,\Delta)$ can be computationally intractable and Variational Bayesian~(VB) is a powerful technique to obtain a tight approximation of it. With a trial distribution $q(\Theta)$, VB solves the following variational optimization:
\begin{equation}
\begin{split}
\mathcal{J}^{VB}(q|X,\Delta) & = \int q(\Theta) \ln\frac{q(\Theta)}{p(X|\Theta,\Delta)p(\Theta|\Delta)} d\Theta,\\
& = \mbox{KL}[q(\Theta)|p(\Theta|X,\Delta)] - \ln p(X|\Delta).
\end{split}
\label{Eq_VBApp}
\end{equation}
The first term in Eq. (\ref{Eq_VBApp}) is the Kullback-Leibler (KL) distance between the trial distribution $q(\Theta)$ and the target posterior distribution $p(\Theta|X,\Delta)$, and the second is constant with respect to $q(\Theta)$. Thus, finding the target posterior distribution $p(\Theta|X,\Delta)$ is equivalent to minimizing the objective function (\ref{Eq_VBApp}) with respect to $q(\Theta)$. In VB approximation, Eq. (\ref{Eq_VBApp}) is minimized over some restricted function space as,
\begin{equation}
\begin{split}
q(\Theta) =& \prod_{j=1}^{J}q(\theta_j).
\end{split}
\label{Eq_VBFact}
\end{equation}
This constrain breaks the entanglement between the latent parameters, and leads to an efficient iterative algorithm. 

With other factors fixed, the variational optimization problem with respect to $q(\theta_j)$ is equivalent to:
\begin{equation}
\begin{split}
&\mathcal{J}^{VB}(q(\theta_j)|X,\Delta) \propto  \int q(\theta_j)\ln {q(\theta_j)}d\theta_j \\
&\hspace{3.5mm} - \int q(\theta_j)\mathbb{E}_{i\neq j}\ln p(\Theta|\Delta) p(X|\Theta,\Delta)d\theta_j,\\
\end{split}
\label{Eq_VBAppV}
\end{equation} 
where the notation $\mathbb{E}_{i \neq j}(\cdot)$ denotes an expectation with respect to the $q$ distributions over all latent parameters $\theta_i$ for $i\neq j$ In this way, the close-form solution of $q(\theta_j)$ satisfies the below condition:
\begin{equation}
\begin{split}
q(\theta_j)\propto \exp[\mathbb{E}_{i\neq j}\ln p(\Theta|\Delta) p(X|\Theta,\Delta)].
\end{split}
\label{Eq_VBAppSol}
\end{equation}
Thus, Eq. (\ref{Eq_VBPoster}) can be solved alternatively by calculating Eq. (\ref{Eq_VBAppSol}) for each parameter.

From the previous discussions, the prior knowledge $p(\Theta|\Delta)$ can be constructed as follows:
\begin{equation}
\begin{split}
p(\Theta|\Delta) &= p(Z|V)p(V|\alpha)p(\tilde{Z}|Z,W)p(Y)\\
&\cdot\prod_{t=1}^{T}p(\mu_t|\mu_0,\Sigma_0)p(A_t|C_A)p(\omega_t|\beta)\\
&\cdot\prod_{k=1}^{K_t}p(u_{t,k}|\varepsilon_0,\Omega_{0})p(\Upsilon_{t,k}|\nu_0, B_{0}),
\end{split}
\label{Eq_VBPrior}
\end{equation}
and 
\begin{equation}
\begin{split}
p(X|\Theta,&\Delta) = \prod_{i=1}^{N} p(x_i|\Theta,\Delta)\\
& = \prod_{i=1}^{N} \mathcal{G}(x_i|A_{z_i}y_{z_i}+\mu_{z_i}+u_{z_i,\tilde{z}_i},\Upsilon_{z_i,\tilde{z}_i}).
\end{split}
\end{equation}
Similar with Eq. (\ref{Eq_VBFact}), we can approximate the posterior distributions over latent parameters of our DDPT model as the following factorized form:
\begin{equation}
\begin{split}
q(\Theta) =& q(Y|Z)q(\tilde{Z}|Z)q(Z)q(V)\cdot\\
&\prod_{t=1}^{T}q(\omega_t)q(\mu_t)q(A_t)\prod_{k=1}^{K_t}q(u_{t,k})q(\Upsilon_{t,k}).
\label{Eq_VBPostFact}
\end{split}
\end{equation}
To simplify our notations, we use $q_i(t)$, $q_i(k|t)$ and $y_{i,t}$ to denote $q(z_i=t)$, $q(\tilde{z}_i=k|z_i =t)$ and $y_{z_i}$ separately. $ \langle f(x) \rangle $ denotes the expectation of $ f(x) $ over $ x $. $ X(d) $ denotes the $ d $th row vector of matrix $ X $. Then we can iteratively solve all the factorized distributions involved in Eq. (\ref{Eq_VBPostFact}), which will be introduced in the following subsections. 

\subsection{Estimation of Low-rank Groups}
The parameters involved in the representation of low-rank groups are $\{Z,V\}\cup\{\mu_t,A_t\}_{t=1}^T$. Based on the prior distributions in Eq. (\ref{Eq_VBPrior}) with Eq. (\ref{Eq_VBAppSol}), we can derive the posterior distributions over $\{Z,V\}\cup\{\mu_t,A_t\}_{t=1}^T$ as follows.

Let $\delta_t = \sum_{i=1}^{N}q_i(t)$, we have
\begin{equation}
\begin{split}
q(V) & \propto\prod_{t=1}^{T}(1-v_t)^{\sum_{j=t+1}^T\delta_j+\alpha-1}v_t^{\delta_t}.
\end{split}
\end{equation}
Thus, the variational posterior of $v_t$ is still beta distributed with $\mbox{Beta}(\alpha_t^1,\alpha_t^2)$ and the parameters are updated with
\begin{equation}
\begin{split}
\alpha_t^1 &= \delta_t+1,\\
\alpha_t^2 &=\alpha+\sum_{j=t+1}^T\delta_j
\end{split}
\end{equation}

Since we have
\begin{equation}
\begin{split}
&\ln\mathcal{G}(x_i|A_ty_{i,t}+\mu_t+u_{t,k},\tau_{t,k})\propto\\
& -0.5\text{Tr}[\Upsilon_{t,k}^{-1}(\mu_{t}\mu_{t}^T+2u_{t,k}\mu_{t}^T+u_{t,k}u_{t,k}^T+x_ix_i^T+y_{i,t}A_tA_t^Ty_{i,t}^T\\
&-2A_{t}y_{i,t}(x_i-\mu_{t}-u_{t,k})^T -2x_i(\mu_{t}+u_{t,k})^T)+\ln |\Upsilon_{t,k}|],
\end{split}
\end{equation}
for updating the posterior distribution of the indicators vector $Z$, we have
\begin{equation}
\begin{split}
q_i(t) & = \frac{r_i(t) }{\sum_{j=1}^{T}r_i(j) },\\
r_i(t) &= \exp(\xi_{t,1}^i+\xi_{t,2}+\xi_{t,3}^i),
\end{split}
\end{equation}
where 
\begin{equation}
\begin{split}
\xi_{t,1}^i &= \langle\ln v_t\rangle =\psi(\alpha_t^1)-\psi(\alpha_t^2),\\
\xi_{t,2}^i & = \sum_{k=1}^{K_t}q_i(k|t)[\psi(\beta_{t,k}^1)-\psi(\beta_{t,k}^2)],\\
\xi_{t,3}^i & = -0.5\sum_{k=1}^{K_t}q_i(k|t) \text{Tr}\{\langle\Upsilon_{t,k}^{-1}\rangle[\langle\mu_{t}\mu_{t}^T\rangle+2\langle u_{t,k}\rangle\langle\mu_{t}^T\rangle+\langle u_{t,k}u_{t,k}^T\rangle\\
&+x_ix_i^T+(\langle y_{i,t}y_{i,t}^T\rangle\langle A_t^TA_t\rangle)-2\langle A_{t}\rangle \langle y_{i,t}\rangle\vartheta_{i,t,k}^T\\
&-2x_i(\langle\mu_{t}\rangle+\langle u_{t,k}\rangle)^T]+\langle\ln|\Upsilon{t,k}|\rangle \}
\end{split}
\end{equation}
Here $\psi(x)$ denotes digamma function defined as $\psi(x) = \frac{d}{dx}\ln\Gamma(x)$ and $ \vartheta_{i,t,k} = x_i-\langle\mu_{t}\rangle-\langle u_{t,k}\rangle $.

For updating the offsetting parameters of each low-rank group $\{\mu_t\}_{t=1}^T$, let $ \iota_{i,t,k} =  x_i-\langle u_{t,k}\rangle-\langle A_{t}\rangle\langle y_{i,t}\rangle $ we have
\begin{equation}
\begin{split}
\ln q(\mu_t)&\propto-0.5\{(\mu_t-\mu_0)^T\Sigma^{-1}_0(\mu_t-\mu_0)\\
&+\sum_{i=1}^{N}\sum_{k=1}^{K_t}q_i(t)q_i(k|t)\mbox{Tr}[\langle \Upsilon_{t,k}^{-1}\rangle[\mu_{t}\mu_{t}^T-2\iota_{i,t,k}\mu_{t}^T]]\}
\end{split}
\end{equation} 
The posterior $q(\mu_t)$ is still a Gaussian distribution with:
\begin{equation}
\begin{split}
\langle{\mu_t}\rangle &= [{\Sigma_0^{-1}}+\sum_{i=1}^{N}\sum_{k=1}^{K_t}q_i(t)q_i(k|t)\langle \Upsilon_{t,k}^{-1}\rangle]^{-1}[\Sigma_0^{-1}\mu_0\\
&+\sum_{i=1}^{N}\sum_{k=1}^{K_t}q_i(t)q_i(k|t)\langle \Upsilon_{t,k}^{-1}\rangle\iota_{i,t,k}],\\
\end{split}
\end{equation} 
and 
\begin{equation}
\begin{split}
\langle{\mu_t\mu_t^T}\rangle = [{\Sigma_0^{-1}}+\sum_{i=1}^{N}\sum_{k=1}^{K_t}q_i(t)q_i(k|t)\langle \Upsilon_{t,k}^{-1}\rangle]^{-1}+\langle{\mu_t}\rangle\langle{\mu_t}\rangle^T.\\
\end{split}
\end{equation}

For updating $v$-th basis $a_{t,v}$ in the dictionary matrix of each group $\{A_t\}_{t=1}^T$, we have:

\begin{equation}
\begin{split}
\ln q(a_{t,v})&\propto-0.5\mbox{Tr}( \frac{1}{c_{a_v}^2}a_{t,v}^T a_{t,v})-0.5\mbox{Tr}(\sum_{i=1}^{N}\sum_{k=1}^{K_t}q_i(t)q_i(k|t) \\
&\cdot \langle\Upsilon_{t,k}^{-1}\rangle[\langle y_{i,t}(v)^2\rangle a_{t,v}a_{t,v}^T-2\langle y_{i,t}(v)\rangle(x_i-\langle\mu_{t}\rangle \\
&-\langle u_{t,k}\rangle-\sum_{j \neq v }^{d}\langle y_{i,t}(j)\rangle \langle a_{t,j}\rangle )a_{t,v}^T]).\\
\end{split}
\end{equation}

Thus, the posterior distribution over $A_t$ satisfies:
\begin{equation}
\begin{split}
\langle A_t \rangle & = [\langle a_{t,1}\rangle,\ldots,\langle a_{t,D}\rangle],\\
\langle a_{t,v}\rangle &= (c_{a_v}^{-2} {\bf I}+\sum_{i=1}^{N}\sum_{k=1}^{K_t}q_i(t)q_i(k|t)\langle \Upsilon_{t,k}^{-1}\rangle{\langle y_{i,t}(v)^2\rangle})^{-1}\\
&\cdot[\sum_{i=1}^{N}\sum_{k=1}^{K_t}q_i(t)q_i(k|t)\langle \Upsilon_{t,k}^{-1}\rangle\langle y_{i,t}(v)\rangle\\
&\cdot(x_i-\langle\mu_{t}\rangle-\langle u_{t,k}\rangle -\sum_{j \neq v }^{d}\langle y_{i,t}(j)\rangle \langle a_{t,j}\rangle ) ], \\
\langle a_{t,v}^Ta_{t,v}\rangle &= \mbox{Tr}(\langle\Upsilon_{t,k}^{-1}\rangle+\sum_{i=1}^{N}\sum_{k=1}^{K_t}q_i(t)q_i(k|t)\langle \Upsilon_{t,k}^{-1}\rangle{\langle y_{i,t}(v)^2\rangle})^{-1}\\
&+\langle a_{t,v}^T\rangle\langle a_{t,v}\rangle.	\\
\end{split}
\end{equation}
$ \langle A_t^TA_t \rangle $ is a $ D \times D $ matrix, and its element $ \langle A_t^TA_t \rangle_{i,j} $ is $\langle a_{t,i}^Ta_{t,j}\rangle $ when $ i = j $ and $ \langle a_{t,i}^T\rangle \langle a_{t,j}\rangle  $ when $ i \neq j $.
\subsection{Estimation of Noise Components}
The parameters involved for modeling complex noise of each low-rank group are $\tilde{Z}\cup\{\omega_t,\{u_{t,k},P_{t,k}\}_{k=1}^{K_t}\}_{t=1}^T$. Based on the prior distributions imposed in Eq. (\ref{Eq_VBPrior}) with Eq. (\ref{Eq_VBAppSol}), we obtain the following posterior distributions over  $\tilde{Z}\cup\{\omega_t,\{u_{t,k},P_{t,k}\}_{k=1}^{K_t}\}_{t=1}^T$.

Let $\lambda_{t,k} =\sum_{i=1}^{N}q_i(t){q_i(k|t)}$, we obtain the posterior distributions over the mixing weights of noise components in each group as follows,
\begin{equation}
\begin{split}
\ln q(\omega_t)&\propto \ln\prod_{k=1}^{K_t} (1-\omega_{t,k})^{\sum_{j=k+1}^{K_t}\lambda_{t,j}+\beta-1} \omega_{t,k}^{\lambda_{t,k}}.\\
\end{split}
\end{equation}
Thus, each $q(\omega_{t,k})$ is also a Beta distribution defined as $\mbox{Beta}(\lambda_{t,k}+1,\sum_{j=k+1}^{K_t}\lambda_{t,j}+\beta)$ with the following parameters,
\begin{equation}
\begin{split}
\beta_{t,k}^1 & = \lambda_{t,k} +1,\\
\beta_{t,k}^2 &= \beta+\sum_{j=k+1}^{K_t}\lambda_{t,j}.
\end{split}
\end{equation}

For updating the indicator vector $\tilde{Z}$, we have
\begin{equation}
\begin{split}
q_i(k|t) & = \frac{\eta_i(k|t)}{\sum_{j=1}^{K_t}\eta_i(j|t)},\\
\eta_i(k|t)& = \varsigma_{t,k,1}^i+\varsigma_{t,k,2}^i,
\end{split}
\end{equation}
where 
\begin{equation}
\begin{split}
\varsigma_{t,k,1}^i &= \psi(\beta_{t,k}^1)-\psi(\beta_{t,k}^2),\\
\varsigma_{t,k,2}^i &=-0.5 \mbox{Tr}\{\langle\Upsilon_{t,k}^{-1}\rangle[\langle\mu_{t}\mu_{t}^T\rangle+2\langle u_{t,k}\rangle\langle\mu_{t}^T\rangle+\langle u_{t,k}u_{t,k}^T\rangle+x_ix_i^T\\
&+(\langle y_{i,t}y_{i,t}^T\rangle\langle A_t^TA_t\rangle)-2\langle A_{t}\rangle \langle y_{i,t}\rangle\vartheta_{i,t,k}^T-2x_i(\langle\mu_{t}\rangle\\
&+\langle u_{t,k}\rangle)^T]+\langle\ln|\Upsilon_{t,k}|\rangle \}.
\end{split}
\end{equation}

 Let $ \iota_{i,t,k}' =  x_i-\langle \mu_t \rangle-\langle A_{t}\rangle\langle y_{i,t}\rangle $. To solve the mean vector $u_{t,k}$ of each noise components in each group, we have
\begin{equation}
\begin{split}
\ln q(u_{t,k})&\propto   -0.5\{(u_{t,k}-\varepsilon_0)^T\Omega_0^{-1}(u_{t,k}-\varepsilon_0)\\
&+\sum_{i=1}^{N}q_i(t)q_i(k|t)\langle \Upsilon_{t,k}^{-1}\rangle[u_{t,k}^Tu_{t,k}-2u_{t,k}^T \iota_{i,t,k}']\}.
\end{split}
\end{equation}
Thus $q(u_{t,k})$ is a Gaussian distribution with:
\begin{equation}
\begin{split}
\langle{u_{t,k}}\rangle &= [\Omega_{0}^{-1}+\sum_{i=1}^{N}q_i(t)q_i(k|t)\langle \Upsilon_{t,k}^{-1}\rangle]^{-1}[\Omega_{0}^{-1}\varepsilon_0\\
&+\sum_{i=1}^{N}q_i(t)q_i(k|t)\langle \Upsilon_{t,k}^{-1}\rangle\iota_{i,t,k}'],
\end{split}
\end{equation}
and 
\begin{equation}
\begin{split}
\langle{u_{t,k}}{u_{t,k}^T}\rangle &= [\Omega_{0}^{-1}+\sum_{i=1}^{N}q_i(t)q_i(k|t)\langle \Upsilon_{t,k}^{-1}\rangle]^{-1}+\langle{u_{t,k}}\rangle\langle u_{t,k}^T\rangle.
\end{split}
\end{equation}

For inferring the posterior distributions over variance of each Gaussian noise component in the group, we have:
\begin{equation}
\begin{split}
\ln q(\Upsilon_{t,k})&\propto -\dfrac{\nu_0 + D + 1}{2}\ln \lvert \Upsilon_{t,k} \rvert-\frac{1}{2}\mbox{Tr}(B_0\Upsilon_{t,k}^{-1})\\
&-0.5\sum_{i=1}^{N}q_i(t)q_i(k|t)\mbox{Tr}[\Upsilon_{t,k}^{-1}(\langle\mu_{t}\mu_{t}^T\rangle+2\langle u_{t,k}\rangle\langle\mu_{t}^T\rangle\\
&+\langle u_{t,k}u_{t,k}^T\rangle+x_ix_i^T+\langle y_{i,t}y_{i,t}^T\rangle\langle A_t^TA_t\rangle-2\vartheta_{i,t,k}^T\langle A_{t}\rangle \langle y_{i,t}\rangle \\
&-2(\langle\mu_{t}\rangle+\langle u_{t,k}\rangle)^Tx_i)+\ln \lvert\Upsilon_{t,k}\rvert].
\end{split}
\end{equation}
The covariance matrix $\Upsilon_{t,k}$ satisfies the following condition:
\begin{equation}
\begin{split}
\langle\Upsilon^{-1}_{t,k}\rangle &= \nu_0'(B_0^{'})^{-1},\\
\langle\ln\lvert \Upsilon_{t,k} \rvert\rangle &= \dfrac{1}{\psi(\nu_0^{'} /2) + D \ln 2 + \ln \lvert (B_0)^{-1} \rvert},
\end{split}
\end{equation}
where 
\begin{equation}
\begin{split}
\nu_0'& = \nu_0 + \lambda_{t,k}\\
B_0' &= B_0+\frac{1}{2}\sum_{i=1}^{N}q_i(t)q_i(k|t)(\langle\mu_{t}\mu_{t}^T\rangle+2\langle u_{t,k}\rangle\langle\mu_{t}^T\rangle\\
&+\langle u_{t,k}u_{t,k}^T\rangle+x_ix_i^T+\langle y_{i,t}y_{i,t}^T\rangle\langle A_t^TA_t\rangle-2\vartheta_{i,t,k}^T\langle A_{t}\rangle \langle y_{i,t}\rangle \\
&-2(\langle\mu_{t}\rangle+\langle u_{t,k}\rangle)^Tx_i)
\end{split}
\end{equation}
\begin{figure*}[t]
	\centering
	\includegraphics[width=1\textwidth]{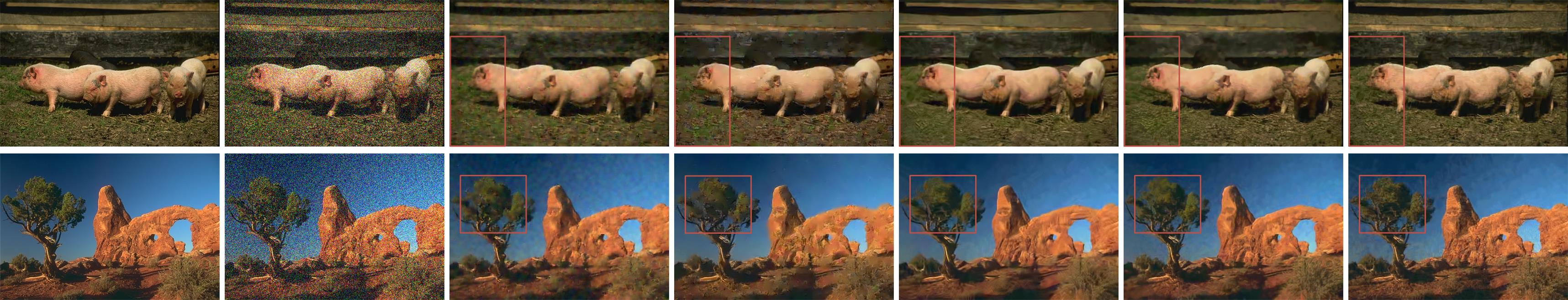}
	\caption{Performance of each algorithm on two images of BSDS500: 66053 and 295087. From left to right: clean image, noisy image with white Gaussian noise~($\sigma = 50$), result with K-SVD, SURE-guided GMM, BM3D, NL-Bayes and our approach. Zoom in to examine the details.}
	\label{fig_bsds}
\end{figure*}
\subsection{Recovering Clean Patch}
To recover the underlying clean patch, we first to calculate the projection $y_{i,t}$ of an observed patch $x_i$ on the $t$-th low-rank group. With Eq. (\ref{Eq_VBAppSol}), we have 
\begin{equation}
\begin{split}
\ln q(Y|Z) &\propto-0.5\sum_{i=1}^{N} [y_i^Ty_i+\sum_{k=1}^{K_t}q_i(k|t)\mbox{Tr}\\
& \cdot[\langle \Upsilon_{t,k}^{-1}\rangle[y_{i,t}\langle A_t^TA_t\rangle y_{i,t}^T-2\langle A_{t}\rangle y_{i,t}\vartheta_{i,t,k}^T]]]
\end{split}
\end{equation}
Thus, $p(y|z_i)$ is still a Gaussian distribution with:
\begin{equation}
\begin{split}
\langle & y_{i,t}\rangle =  \\
&[{\bf I}+\sum_{k=1}^{K_t}q_i(k|t)\langle \Upsilon_{t,k}^{-1}\rangle A_t^TA_t]^{-1}[\sum_{k=1}^{K_t}q_i(k|t)\langle \Upsilon_{t,k}^{-1}\rangle\vartheta_{i,t,k}^T\langle A_{t}\rangle],
\end{split}
\end{equation}
and 
\begin{equation}
\begin{split}
\langle y_{i,t}y_{i,t}^T\rangle &=  [{\bf I}+\sum_{k=1}^{K_t}q_i(k|t)\langle \Upsilon_{t,k}^{-1}\rangle \langle A_t^TA_t\rangle]^{-1}+\langle y_{i,t}\rangle\langle y_{i,t}^T\rangle ,
\end{split}
\end{equation}

Iteratively performing (13)-(34), the algorithm will converge to at least a local minimum. And a clean patch represented by $t$-th group can be well-recovered with
\begin{equation}
\label{Eq_wiener}
\begin{split}
\hat{x}_i(t) = \langle A_t\rangle\langle y_{i,t}\rangle+\langle\mu_t\rangle.
\end{split}
\end{equation}
The above recovery method can also be interpreted as a novel \textit{Variational Bayesian Wiener Filtering}~\cite{wiener1949extrapolation} which can be utilized to recover signals contaminated unknown MoG noise. It can be utilized to generalize other denoising algorithm based on wiener filtering to blind image denoising including \cite{dabov2007image,buades2005non}. 

The latent variable $ t $ corresponding to the subspace that $ x_i $ belongs to, can be obtained with
\begin{align}
\arg \max_t  q_i(t) . 
\end{align}
With the recovered patches, we aggregate them to obtain clean image following the algorithm in \cite{dabov2007image}.

\section{Experiments}
\label{sec_Experiment}
In this section, we extensively evaluate our approach by comparing it with a number of state-of-the-art algorithms. The experimental results show that our method can handle niosy images with large variety of noise models in real applications with superior performance compared with previous works.
\subsection{Parameter Setting}
Throughout this work, we use the same parameter as follows~(without tuning): the patch size is set to be 8x8. We extract patches from each input image by the same algorithm in conventional patch based denoising algorithms~\cite{buades2005non,dabov2007image}.  For the DDPT, we set the top-level DP concentration parameters to $ \alpha = 3 $ and the second-level DP concentration parameter to $ \beta =  10^{-3} $. For other hyper-parameters, we set $ \mu_0, \varepsilon_0$ be $0$; $ \nu_0$ be 64~(dimension of patch vector); and $ B_0, C_a, \Sigma_0, \Omega_0 $ be $ I $. With these simple settings, our approach performs stably well on all the following experiments. 
\begin{table*}
	\centering
	\caption{Performance of different approaches on noisy images with homogeneous white Gaussian noise with different deviation.}
	\label{Table_I}
	\begin{tabular}{ |l|l|l|l|l|l|l|l|l|l|l|l|l| } 
		\hline 
		& $\sigma$	& 10			& 20			& 30			& 40			& 50			& 60			& 70			& 80			& 90			& 100  \\ 
		\hline 
		Dataset						&	Method	& \multicolumn{10}{ |c| }{PSNR} \\ 
		\hline 
		\multirow{5}{*}{TID2008} 	& K-SVD 	& 34.74			& 30.87			& 28.97			& 27.89			& 26.04			& 24.92 		& 24.45			& 23.80 		& 23.33			& 22.84			\\ 
		& SURE-GMM 	& 34.79			& 31.22			& 29.23			& 27.91 		& 26.53			& 24.89			& 24.78			& 23.81			& 23.70			& 23.04			\\
		& BM3D 		& \textbf{35.17}& 31.46			& 29.28			& 28.02			& 26.66			& 25.39			& 25.14			& 24.66 		& 24.11			& 23.26			\\ 
		& NL-Bayes 	& 35.06			& \textbf{31.51}& \textbf{29.31}& 28.04			& 26.67			& \textbf{25.43}& 25.11			& 24.51 		& 23.84			& 23.30			\\
		& Ours		& 35.11			& 31.43			& 29.26			& \textbf{28.07}& \textbf{26.74}& 25.38 		& \textbf{25.19}& \textbf{24.71}& \textbf{24.21}& \textbf{23.32}\\ 
		\hline 
		\multirow{5}{*}{BSDS500} 	& K-SVD 	& 34.24			& 30.69			& 28.66			& 27.69			& 25.93			& 25.10 		& 24.22			& 23.46 		& 23.12			& 22.67			\\ 
		& SURE-GMM 	& 34.22			& 31.01			& 29.02			& 27.68 		& 26.23			& 25.22			& 24.56			& 23.58			& 23.53			& 22.97			\\
		& BM3D 		& \textbf{34.72}& \textbf{31.19}& 29.11			& 27.51			& 26.56			& \textbf{25.68}& 24.92			& 24.39 		& 23.97			& 23.02			\\ 
		& NL-Bayes 	& 34.69			& 31.09			& \textbf{29.09}& 27.53			& 26.48			& 25.66			& 24.79			& 24.41 		& 23.68			& 23.07			\\
		& Ours		& 34.71			& 31.11			& 29.06			& \textbf{27.85}& \textbf{26.61}& 25.64 		& \textbf{24.94}& \textbf{24.47}& \textbf{24.00}& \textbf{23.13}\\  
		\hline 
		\multicolumn{2}{|c|}{}		& \multicolumn{10}{ |c| }{SSIM} \\ 
		\hline 
		\multirow{5}{*}{TID2008} 	& K-SVD 	& 0.947			& 0.930			& 0.900			& 0.861			& 0.841			& 0.817 		& 0.798			& 0.775 		& 0.764			& 0.743			\\ 
		& SURE-GMM 	& 0.944			& 0.935			& 0.904			& 0.868 		& 0.848			& 0.818			& 0.807			& 0.787			& 0.770			& 0.747			\\
		& BM3D 		& \textbf{0.968}& \textbf{0.938}& 0.912			& 0.874			& 0.856			& 0.828			& 0.816			& 0.800 		& 0.781			& 0.764			\\ 
		& NL-Bayes 	& 0.962			& 0.934			& \textbf{0.916}& 0.875			& 0.860			& 0.827			& \textbf{0.818}& 0.794 		& 0.779			& 0.761			\\
		& Ours		& 0.958			& 0.933			& 0.912			& \textbf{0.877}& \textbf{0.862}& \textbf{0.829}& 0.816			& \textbf{0.802}& \textbf{0.788}& \textbf{0.770}\\ 
		\hline 
		\multirow{5}{*}{BSDS500} 	& K-SVD 	& 0.938			& 0.925			& 0.897			& 0.856			& 0.830			& 0.822 		& 0.790			& 0.766 		& 0.762			& 0.741			\\ 
		& SURE-GMM 	& 0.939			& 0.923			& 0.899			& 0.859 		& 0.833			& 0.827			& 0.793			& 0.774			& 0.768			& 0.746			\\
		& BM3D 		& \textbf{0.961}& \textbf{0.931}& 0.907			& \textbf{0.867}& 0.848			& 0.835			& 0.806			& 0.792 		& 0.777			& 0.763			\\ 
		& NL-Bayes 	& 0.953			& 0.926			& 0.908			& 0.866			& 0.849			& 0.834			& 0.810			& 0.793 		& 0.772			& 0.764			\\
		& Ours		& 0.956			& 0.929			& \textbf{0.910}& 0.866			& \textbf{0.851}& \textbf{0.836}& \textbf{0.811}& \textbf{0.798}& \textbf{0.779}& \textbf{0.767}\\   
		\hline
	\end{tabular}
\end{table*}
We initialize the parameters of our model as follows. We first perform k-means++ algorithm~\cite{arthur2007k} to divide the patches into groups and initialize $ \mu_t $ with the mean of each group. The number of components is set to be 30. For each group $ t $, we initialize $ \mu_t $ by getting the average of the component. Then we perform singular value decomposition~(SVD)~\cite{bishop2006pattern} to initialize $ A_t $. Then we can get the projection of each patch on the subspace within a component. For each group $ t $, we calculate the residual of each patch from its projection and perform k-means++~($ k = 10 $) again and use its result as the initialization of $ \boldsymbol{\phi_t} $. We use the above scheme for parameter initialization for all the experiments and find it robust for different noisy images. 
\subsection{Homogeneous White Gaussian Noise}
\label{normal_noise}
Our method can be used to handle noisy images with different noise models. We first test the performance of our method on images contaminated by homogeneous white Gaussian noise which is a general assumption of noise for traditional non-blind denoising algorithms. Here, we compare our method with several state-of-the-art non-blind denoising algorithms designed for homogeneous white Gaussian noise: K-SVD\footnote{\url{http://www.cs.technion.ac.il/~elad/software/}}~\cite{elad2006image}, SURE-guided GMM\footnote{\url{http://www.ipol.im/pub/art/2013/52/}}~\cite{wang2013sure}, BM3D\footnote{\url{http://www.cs.tut.fi/~foi/GCF-BM3D/index.html#ref_software}}~\cite{dabov2007image} and NL-Bayes\footnote{\url{http://www.ipol.im/pub/art/2013/16/}}~\cite{lebrun2013nonlocal}. We test all the algorithms on synthesis noisy images which are produced by adding homogeneous white Gaussian noise with different deviation to images from two benchmark datasets of clean images: TID2008\footnote{\url{http://www.computervisiononline.com/dataset/tid2008-tampere-image-database-2008}}~\cite{ponomarenko2009tid2008} and BSDS500\footnote{\url{http://www.eecs.berkeley.edu/Research/Projects/CS/vision/grouping/resources.html}}~\cite{MartinFTM01}. As all the competitive algorithms are non-blind, we provide the true noise intensity of each noisy image as input of these algorithms. For our blind approach, we let the algorithm to inference the noise intensity automatically. To evaluate the performance of each algorithm, we introduce two measurements: peak signal-to-noise ratio~(PSNR) and the structural similarity index~(SSIM)~\cite{wang2004image} which are used to measure the similarity between the denoised image and the clean one. A larger value of either PSNR or SSIM indicates that the performance is better.

Table.~\ref{Table_I} illustrates the numerical performance of all the algorithms on synthesis images contaminated by noise with intensity from 10 to 100. It can be observed that, with both measurements, our algorithm achieves competitive performance compared with state-of-the-art algorithms especially when the images are contaminated by large noise intensities~($ \sigma > 40 $). Fig.~\ref{fig_bsds} shows the denoising results of two images from BSDS500 with noise intensity $ \sigma = 50 $ for visualization. With the modeling of clean patches, our algorithm can efficiently eliminate noise while reserving the detailed features especially when the noise intensity is large. Specifically, our method is the only one to reserve the patterns of grass in image 66053 and tree in 295087 while others tend to smooth these features.

\begin{figure*}[]
	\centering
	\includegraphics[width=0.84\textwidth]{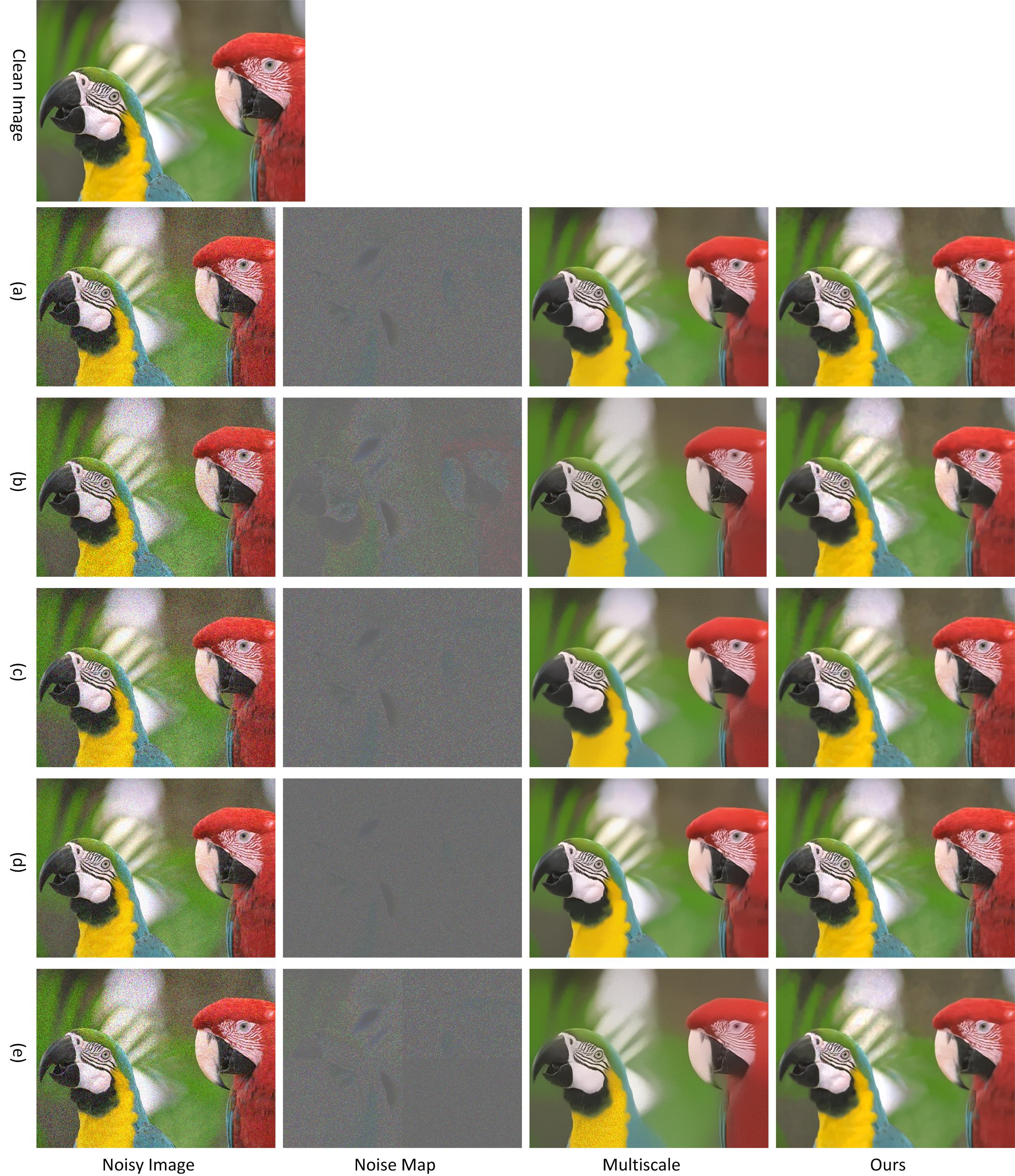}
	\caption{Comparison of \cite{lebrun2014multiscale} and ours on I23 from TID2008 contaminated by (a) homogeneous white Gaussian noise with $ \sigma = 30 $, PSNR: multiscale = 28.9064, ours = \textbf{32.5856}; (b) heterogeneous noise with $ b = 4 $, PSNR: multiscale = 25.2097, ours = \textbf{30.7666}; (c) Laplace noise with $ \sigma = 30 $, PSNR: multiscale = 28.3428, ours = \textbf{32.4594}; (d) uniform noise with $ a = 30 $, PSNR: multiscale = 32.1877, ours = \textbf{35.2331}; (e) combined noise, PSNR: multiscale = 25.6925, ours = \textbf{30.8966}.}
	\label{fig_I23}
\end{figure*}

%
\begin{figure}
	\centering
	\includegraphics[width=0.5\textwidth]{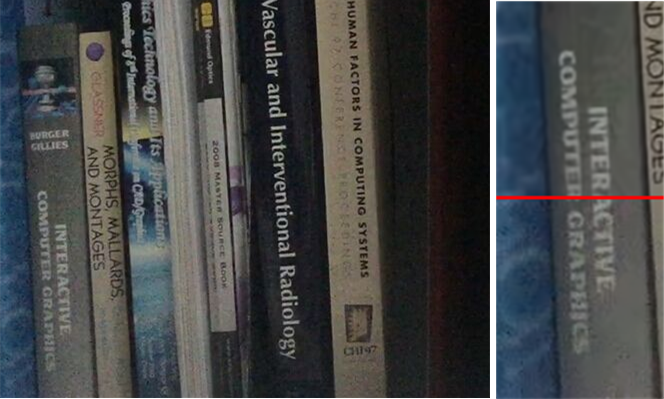}
	\caption{A real noisy image captured by CCD camera and a crop of the denoising results of it. The result above the red line is ours and that blow it is \cite{lebrun2014multiscale}.}
	\label{fig_ccd1}
\end{figure}
\subsection{Arbitrary Noise}
\label{Diff_noise}
For practical image denoising, the underlying noise model of real noisy images can be different from homogeneous white Gaussian noise and a blind denoising algorithm should have the capability to handle images contaminated by different noise. In this section, we propose experiments on synthesis data to show that our approach can handle images contaminated by different types of noise even the noise model is not provided. The noisy images are produced by adding the following types of zero-mean noise to clean images of TID2008 and BSDS500:
\begin{itemize}
	\item Homogeneous white Gaussian noise with $ \sigma = 15, 30, 45 $.
	\item Heterogeneous Gaussian noise with $ \sigma = \frac{1}{b} x_{i,j} $, where $ x_{i,j} $ is the intensity of pixel in position $ (i, j) $ with $ b = 3, 4, 5 $ resulting to intensity-dependent noise.
	\item Laplace noise with $ \sigma = 15, 30, 45 $.
	\item Uniform noise of $ [-a, a] $ with $ a = 15, 30, 45 $.
	\item The combination of the above 4 types of noise: we averagely divide an image into 4 parts: left up, right up, left down, right down. The four parts are contaminated by heterogeneous Gaussian noise with $ b = 4 $, Laplace noise with $ \sigma = 30 $, white Gaussian noise with $ \sigma = 30 $ and uniform noise with $ a = 30 $ respectively resulting to position dependent noise. 
\end{itemize} 
\begin{figure}
	\centering
	\includegraphics[width=0.5\textwidth]{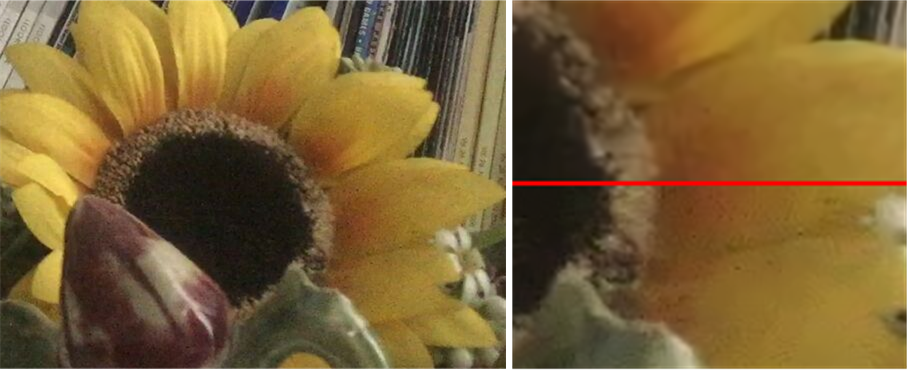}
	\caption{A real noisy image captured by CCD camera and a crop of the denoising results of it. The result above the red line is ours and that blow it is \cite{lebrun2014multiscale}.}
	\label{fig_ccd2}
\end{figure}
We apply our methods to the noisy images contaminated by the above types of noise with comparison of the state-of-the-art blind image denoising approach which is also claimed to be able to handle heterogeneous non-Gaussian noise, multiscale noise clinic\footnote{\url{http://www.ipol.im/pub/art/2015/125/}}~\cite{lebrun2014multiscale}~(note that we only compare with this algorithm as it is the state-of-the-art algorithm in this area as shown in \cite{lebrun2014multiscale} and it is the only blind image denoising approach with code online available). The multiscale approach has a free parameter, e.g., the number of scales, and the researchers claimed that it can handle all images in the range from 2 to 5. For fair comparison, for each noisy image, we run the multiscale approach with number of scales from 2 to 5 and select the best one as their denoising result. For our method, the parameters are set to be fixed as introduced before. Table~\ref{Table_II} illustrates the numerical performance of the two algorithms. It can be observed that our method is better than the multiscale one with respect to both PSNR and SSIM with a significant margin for all types of noise with different parameters. Fig.~\ref{fig_I23} shows an example of the denoising results for image I23\footnote{The small patterns on Fig.~\ref{fig_I23}(a)(c)(d) are formed as the image intensity is bounded by [0,255]. When the value of a noisy pixel is smaller than 0 or larger than 255, it well be regularized.} in TID2008. It can be observed that the multiscale approach tends to smooth the detailed features~(the feathers of the birds). Moreover, some noise is not eliminated completely when the noise is heterogeneous~(e.g., the yellow part of Fig.~\ref{fig_I23}(e)). This is caused by the less of generality of the noise model in~\cite{lebrun2014multiscale}. As a comparison, due to the flexibility of MoG, our approach can well eliminate the noise and the features are well preserved. It shows that our approach is general enough to handle noisy images contaminated by different noise model. 
\begin{landscape}
{
	\begin{table}
	\centering
	\caption{Performance of the multiscale approach and ours on images contaminated by five different types of noise.}
	\label{Table_II}
	\begin{tabular}{ |l|l|l|l|l|l|l|l|l|l|l|l|l|l|l| }
		\hline
		&											& \multicolumn{3}{|c|}{Gaussian}					& \multicolumn{3}{|c|}{Heterogeneous}				& \multicolumn{3}{|c|}{Laplace}		& \multicolumn{3}{|c|}{Uniform}		& Comb.	\\
		\hline
		&											& 15			& 30			& 45				& 3 	& 4 	& 5								& 15 	& 30	& 45				& 15 	& 30 	& 45				&				\\
		\hline
		Dataset						& Method		&	\multicolumn{13}{|c|}{PSNR}	\\
		\hline
		\multirow{2}{*}{TID2008} 	& Multiscale 	& 30.27 		& 27.02 		& 25.14				& 24.07			& 25.54			& 26.16			& 29.78			& 26.87			& 24.97			& 			34.22			       						& 31.77			& 28.97			& 25.12\\
		& Ours			& \textbf{33.22}& \textbf{29.26}& \textbf{27.22} 	& \textbf{27.64}& \textbf{28.94}& \textbf{30.02}& \textbf{33.11}& \textbf{29.07}& \textbf{27.14}& \textbf{36.23}	& \textbf{35.27}& \textbf{32.16}& \textbf{29.87}\\
		\hline
		\multirow{2}{*}{BSDS500} 	& Multiscale 	& 29.47 		& 26.83 		& 25.00				& 23.87			& 25.12			& 25.47			& 29.63			& 26.53			& 24.29			& 			34.22			       						& 30.54			& 28.24			& 24.82\\
		& Ours			& \textbf{32.87}& \textbf{28.41}& \textbf{29.93} 	& \textbf{27.13}& \textbf{28.29}& \textbf{30.07}& \textbf{32.76}& \textbf{28.55}& \textbf{26.87}& \textbf{35.73}	& \textbf{34.84}& \textbf{31.06}& \textbf{29.38}\\
		\hline
		\multicolumn{2}{|c|}{}						&	\multicolumn{13}{|c|}{SSIM}	\\
		\hline
		\multirow{2}{*}{TID2008} 	& Multiscale 	& 0.913 		& 0.871 		& 0.831			& 0.782			& 0.841			& 0.836			& 0.918			& 0.868			& 0.831				& 0.952						& 0.943			& 0.922			& 0.822\\
		& Ours			& \textbf{0.952}& \textbf{0.912}& \textbf{0.867}& \textbf{0.870}& \textbf{0.899}& \textbf{0.918}& \textbf{0.949}& \textbf{0.901}& \textbf{0.863}	& \textbf{0.977}& \textbf{0.961}& \textbf{0.930} & \textbf{0.903}\\
		\hline
		\multirow{2}{*}{BSDS500} 	& Multiscale 	& 0.912 		& 0.868 		& 0.826			& 0.778			& 0.840			& 0.829			& 0.912			& 0.859			& 0.836				& 0.952						& 0.944			& 0.909			& 0.817\\
		& Ours			& \textbf{0.948}& \textbf{0.910}& \textbf{0.862}& \textbf{0.867}& \textbf{0.893}& \textbf{0.913}& \textbf{0.942}& \textbf{0.897}& \textbf{0.857}	& \textbf{0.970}& \textbf{0.953}& \textbf{0.921} & \textbf{0.904}\\
		\hline
	\end{tabular}
\end{table}
}
\end{landscape}
\begin{figure}
	\centering
	\includegraphics[width=0.9\textwidth]{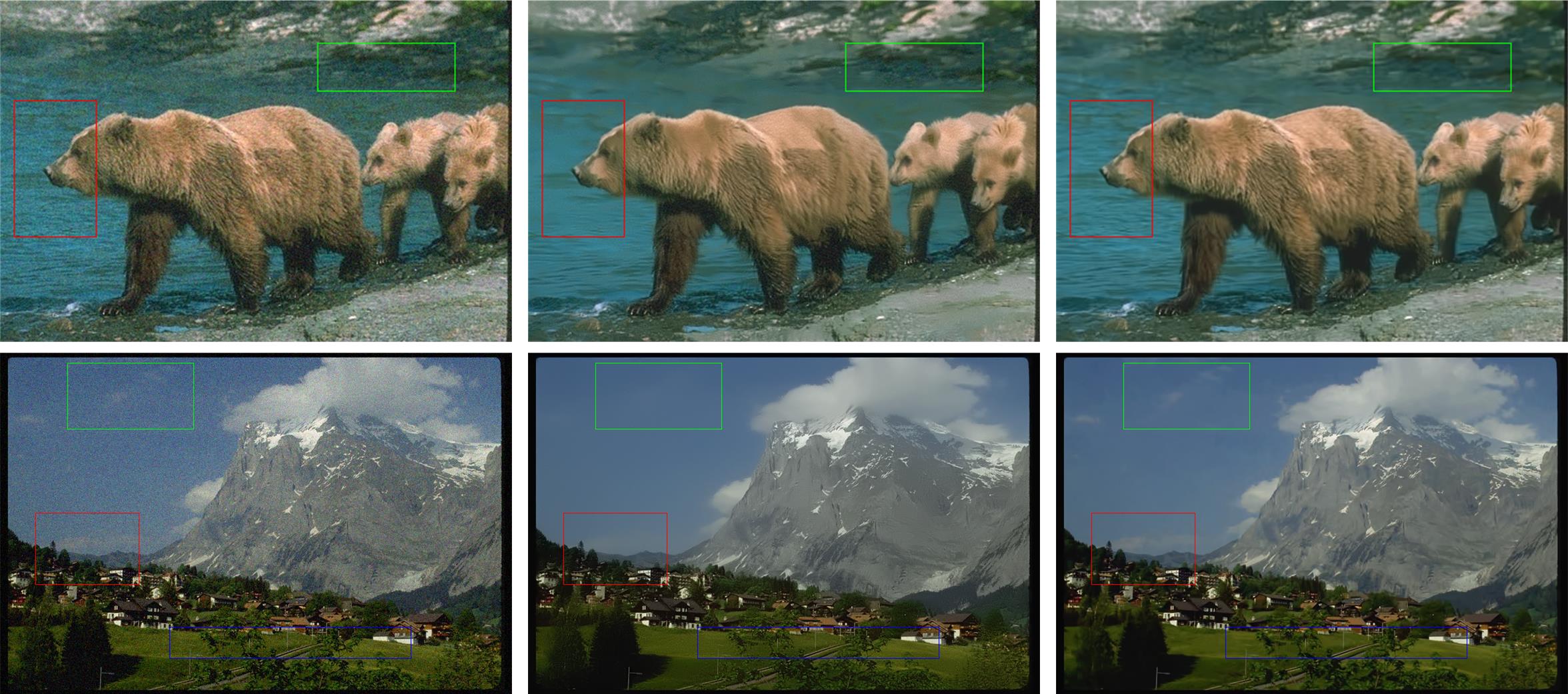}
	\caption{From left to right: observed noisy images, the denoising results of \cite{lebrun2014multiscale} and ours. Zoom in to examine the details.}
	\label{fig_others}
\end{figure}
\subsection{Real Noisy Images}
\label{real_image}
To show the efficiency of our approach in real-world problems, we further demonstrate the performance of our method on handling real noisy images with a comparison of the method in \cite{lebrun2014multiscale}. These algorithm are applied on images with various noise models. We first test our method using pictures taken by CCD cameras with remarkable noise. Fig.~\ref{fig_ccd1} and \ref{fig_ccd2} were taken by Nikon D5200~(ISO 6400, exposure time 1/20s and aperture f/5). We apply both algorithms on them and it can be seen that ours can better eliminate noise without introducing artifacts. Fig.~\ref{fig_others} shows the performance of another two example noisy images, `bear' and `postcard'~\cite{lebrun2014multiscale} whose noise is relatively ``normal''. It can be observed that, comparing with the result of \cite{lebrun2014multiscale}, ours can better eliminate the noise and preserve the features. Specifically, the features marked by green and red boxes are smoothed by the multiscale approach while well reserved by ours. Meanwhile, within the blue box, the noise is not well handled by the multiscale approach while well eliminated by ours.   

\begin{figure}
	\centering
	\includegraphics[width=0.9\textwidth]{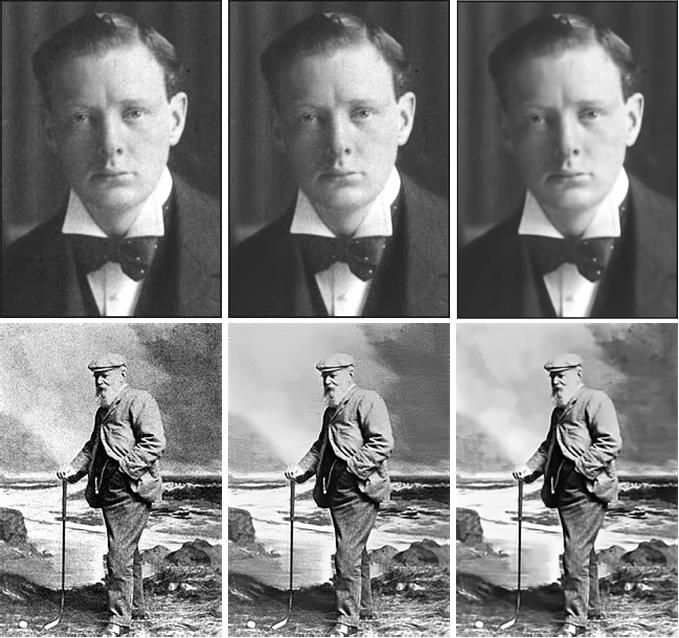}
	\caption{From left to right: old photos of young Winston Churchill and Tomas Morris, the denoising results of \cite{lebrun2014multiscale} and ours. }
	\label{fig_churchill}
\end{figure}

\begin{figure}[t]
	\centering
	\includegraphics[width=0.6\textwidth]{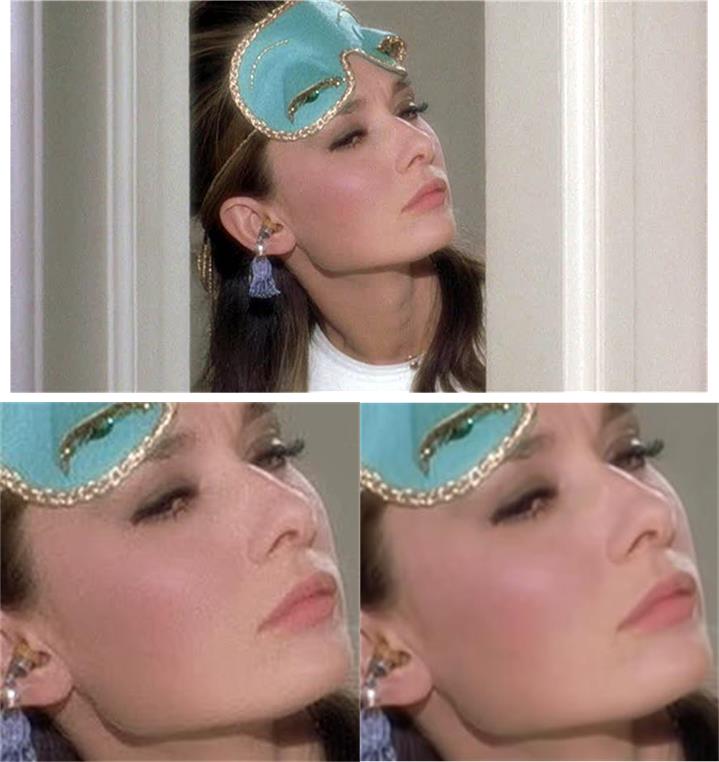}
	\caption{A screen-shot of the old movie `Breakfast at Tiffany's'. Left is a crop of the denoising results of \cite{lebrun2014multiscale} and ours is at right.}
	\label{fig_sleepmask}
\end{figure}

We further apply the algorithms on the recovery of scanned old photographs and screen-shots of old movies. The noise of these images is generally with large grain and altered by further processing including scanning and JPEG encoding. Fig.~\ref{fig_churchill} and \ref{fig_sleepmask} show several examples of the results obtained by the two algorithms over this kind of noise. It can be observed that ours can eliminate this kind of noise efficiently. It shows that our noise model can better handle large and even structural noise on real images. In comparison, \cite{lebrun2014multiscale} can also eliminate some noise but not completely. Fig.~\ref{fig_churchill} shows the obtained results of another two old photographs of David Hilbert and Tom Morris, which were downloaded from their Wikipedia profiles. Our approach achieves better performance on all the images. In contrast, \cite{lebrun2014multiscale} can not eliminate the noise completely. Same phenomena appears in the denoising results for other images as well. It shows the efficiency of our algorithm on handling real noisy images even when the noise is complex.
\section{Conclusion}
\label{sec_Conclu}
We proposed a learning-based approach to automatically recover the clean image from the observed noisy one. The noise model is unknown and modeled with a MoG. The clean patches are assumed to lie in several local subspaces. We built a two-layer structural mixture model for noisy patches and treat the clean ones as latent variables. To build the model, we proposed the dependent Dirichlet process tree as prior, which is a novel nonparametric prior that introduces a mechanism of parameters sharing among mixtures. A variational inference algorithm was proposed accordingly to estimate both the model and the clean patches as latent variables. Extensive experiments were conducted to test the performance of our approach on images contaminated by different noise. Our method achieved the best performance among competitive algorithms, preserving detailed features and eliminating noise with different models. These features make our method a better candidate in handling real-world image denoising tasks. 

We point out several directions for future works based on the proposed work:
\begin{itemize}
	\item[1.] We modeled the complex on image with MoG, which is the first time on image noise modeling. This idea can be used to generate other previous non-blind image denoising approaches and develop new blind image denoising algorithms for images with unknown noise.
	\item[2.] We developed the DDPT model as a prior for our noisy patch model. This prior introduces the mechanisms of parameter sharing among mixture components. Though we focused on blind image denoising in this paper, the proposed model can be utilized in different applications, e.g., fitting topic models~\cite{blei2003latent}, discovering taxonomies of images~\cite{bart2008unsupervised}.  
\end{itemize}

\bibliographystyle{IEEEtran}
\bibliography{ref}
%




\end{document}